\documentclass[review]{elsarticle}

\usepackage{amsmath}
\usepackage{amssymb}
\usepackage{commath}
\usepackage{comment}
\usepackage{graphicx}
\usepackage{booktabs}
\usepackage{multirow}
\usepackage{colortbl}
\usepackage{tabularx}

\usepackage{subfig}
\usepackage{lineno,hyperref}
\modulolinenumbers[5]

\biboptions{authoryear}

\begin{document}

\begin{frontmatter}

\title{PCA-Based Missing Information Imputation for Real-Time Crash Likelihood Prediction Under Imbalanced Data}

%%% Group authors per affiliation:

\author[mymainaddress]{Jintao~Ke}
\author[mysecondaryaddress]{Shuaichao Zhang}
\author[mymainaddress]{Hai Yang}

\author[mysecondaryaddress]{Xiqun~(Michael)~Chen\corref{mycorrespondingauthor}}
\cortext[mycorrespondingauthor]{Corresponding author}
\ead{chenxiqun@zju.edu.cn}

\address[mymainaddress]{Department of Civil and Environmental Engineering, Hong Kong University of Science and Technology, Clear Water Bay, Kowloon, Hong Kong, China}
\address[mysecondaryaddress]{College of Civil Engineering and Architecture, Zhejiang University, Hangzhou 310058, China}

\begin{abstract}

The real-time crash likelihood prediction has been an important research topic. Various classifiers, such as support vector machine (SVM) and tree-based boosting algorithms, have been proposed in traffic safety studies. However, few research focuses on the missing data imputation in real-time crash likelihood prediction, although missing values are commonly observed due to breakdown of sensors or external interference. Besides, classifying imbalanced data is also a difficult problem in real-time crash likelihood prediction, since it is hard to distinguish crash-prone cases from non-crash cases which compose the majority of the observed samples. In this paper, principal component analysis (PCA) based approaches, including LS-PCA, PPCA, and VBPCA, are employed for imputing missing values, while two kinds of solutions are developed to solve the problem in imbalanced data. The results show that PPCA and VBPCA not only outperform LS-PCA and other imputation methods (including mean imputation and $k$-means clustering imputation), in terms of the root mean square error (RMSE), but also help the classifiers achieve better predictive performance. The two solutions, i.e., cost-sensitive learning and synthetic minority oversampling technique (SMOTE), help improve the sensitivity by adjusting the classifiers to pay more attention to the minority class.

\end{abstract}


\end{frontmatter}

% \linenumbers

\section{Introduction}
Prediction of traffic crash has been a major research topic in transportation safety studies. Crashes, especially on urban expressways, can trigger heavy traffic congestions, impose huge external costs, and reduce the level of service of transportation infrastructures. Therefore, the accurate and reliable prediction of crash risks is critical to the success of proactive safety management strategies on urban expressways.

There have been fruitful studies in the domain of the real-time crash likelihood estimation \citep{abdel2006calibrating,abdel2007crash,abdel2008assessing,ahmed2012viability}. It has been reported that crash occurrence was affected by four major factors: real-time traffic state, drivers' behavior, environment factors, and road geometry \citep{ahmed2013data}. Traditional devices utilized in detecting real-time traffic states are mainly intrusive, e.g., loop detectors. Recently, more non-intrusive traffic detection devices are in use due to their easiness of installation, maintenance, accuracy, and affordable costs. For example, Remote Traffic Microwave Sensors (RTMS) and Automatic Vehicle Identification (AVI) devices provide access to real-time traffic data from multiple sources. In field applications, RTMS simultaneously provide real-time data of flow, time occupancy, and speed.

Despite RTMS or other detectors (e.g., AVI devices and loop detectors) have been widely used and successfully applied in traffic operations including the real-time crash likelihood estimation, the problem of missing information has generated trouble for researchers and traffic operators for years \citep{turner2000archived,chen2002quality,chen2003detecting,smith2003exploring,rajagopal2007health}. Dynamic traffic flow data in the intelligence transportation systems unavoidably face with the missing data issue mainly due to the detector failure and lossy communication systems \citep{asif2016matrix}. According to \cite{ahmed2012viability}, loop detectors have a failure that ranges between 24\% and 29\%. It is reported that 5\% of traffic data are lost at hundreds of detection points within PeMS traffic flow database \citep{li2013efficient}. In some extreme cases, the missing percentage can reach 90\%, which has become a critical issue for traffic management \citep{tan2013tensor}. On the other hand, dynamic traffic flow data serve as one of the most important components of the features in real-time crash likelihood estimation. Therefore, the issue of missing data is not negligible, since it may greatly affect the predictive performance of the models. However, in the crash likelihood estimation literature, most existing methods and algorithms are developed under the assumption of complete data (the samples with missing data are simply deleted). Although there were some studies on the missing data imputation for traffic flow measurements \citep{li2014missing,li2013efficient,li2013comparison,li2014traffic}, the patterns and characteristics of traffic crash data are different from those of traffic flow data. One main distinction is that traffic crash data do not have periodicity and tendency, which are important properties in traffic flow data. Few studies have been implemented to seek out the suitable imputation approaches for imputing missing crash data. Furthermore, the behavior, robustness, and properties of the predictive models under highly missing data have not been fully understood in real-time crash likelihood estimation.

Another problem, which has attracted little attention in real-time crash likelihood prediction, is the imbalanced issue of the field measurements. It is a typical imbalanced classification problem because the number of crash occurrence samples is much smaller than the number of non-crash samples. In this paper, two kinds of solutions are employed to solve the imbalanced issue, including the cost-sensitive learning at the algorithmic level and the synthetic minority oversampling technique (SMOTE) at the data level. An imbalanced dataset may produce biased classification results towards the majority class, the reasons for which include: (I) the classifiers regard the costs of misclassifying positive or negative samples as the same; (II) the objective of the algorithm tends to reduce the total errors on which the minority class has few influences.

To bridge the two research gaps in real-time crash likelihood estimation, the paper aims to incorporate the two important components, i.e., missing data imputation, and solutions to the imbalanced issue in real-time crash likelihood estimation. Firstly, various imputation approaches are examined and compared in the imputation of crash data, which have different patterns and distributions from traditional traffic flow data. To understand characteristics of the missing patterns in crash data and find the most suitable imputation approach are important for the field application of real-time crash likelihood estimation. Secondly, different classifiers' tolerance and robustness to missing data are studied, especially when the missing ratio is high. Those classifiers with the low tolerance to missing data should be avoided under the scenario of highly missing data. Thirdly, two solutions to imbalanced issues are adopted and compared in real-time crash likelihood estimation.

The rest of the paper is organized as follows: Section 2 reviews related studies. Section 3 presents the methodologies in the domain of real-time crash likelihood estimation, missing data imputation, and solutions to imbalanced data. Section 4 demonstrates the numerical test of an urban expressway in Hangzhou, China, and presents the results of sensitivity analyses. Finally, Section 5 concludes the whole paper, summarizes the interesting findings of this paper, and outlooks future research.

\section{Literature Review}

Since the beginning of the last decade, researchers have employed a board range of machine learning algorithms in real-time crash likelihood estimation, mainly utilizing factors extracted from traffic dynamics, such as flow, occupancy, and speed. In the early studies of this domain, empirical models based on crash precursors \citep{lee2003real}, nonparametric Bayesian model \citep{oh2005real}, were developed to predict the potential crash occurrence. \cite{abdel2006calibrating} designed a matched case-control Logit model, which integrated traffic flow data and weather data into real-time crash likelihood estimation. Support vector machine, a classical machine learning algorithm, was widely used to predict the crash occurrence due to its robustness on small datasets \citep{yu2013utilizing,sun2014use}. Tree-based ensemble methods, such as the stochastic gradient boosting (SGB) and AdaBoost, also demonstrated strong ability in enhancing reliability of the real-time risk assessment \citep{ahmed2013application}. Motivated by the fact that a large number of explanatory variables might induce overfitting issues, random forest was widely used for selecting important factors \citep{saha2015random}. \cite{kwak2016predicting} used the conditional logistic regression analysis to remove confounding factors, and developed separate models to predict crash occurrence with the genetic programming technique. Recently, more and more researchers employed Bayesian approaches, such as the hybrid latent class analysis (LCA), dynamic Bayesian network (DBN), and semi-parametric based model, to further reveal the unobserved heterogeneity among crashes \citep{sun2015dynamic,yu2016crash,yu2013multi,yu2014using,xu2014using}. \citet{roshandel2015impact} compared the accumulated approaches in this area and summarized the current knowledge, challenges and opportunities of assessing the impact of traffic factors on the crash occurrence.

Apart from developing state-of-the-art approaches for real-time crash risk assessment, researchers have made various attempts to seek for more relevant explanatory variables. \citet{el2014investigation} investigated the impact of the sudden extreme snow or rain variation on the crash type, using full Bayesian multivariate Poisson log-normal models. \citet{yu2014utilizing} proposed a hierarchical logistic regression model to predict crash likelihood, with multi-source information, including traffic, weather, and roadway geometric factors.

On the other hand, the difference existing in the types of crash has also attracted attention from many researchers. \cite{sun2016analysis} built separate models for non-congested-flow crashes and congested-flow crashes and compared their safety factors. Although most of the literatures defined dependent variables as a binary variable (crash occurrence or not), some researchers made attempts to predict crash likelihood at various levels of severity \citep{xu2013predicting}. There were some studies concentrating on assessing the risk of rear-end crash, which was regarded as the most severe crash type \citep{weng2014rear,lao2014generalized,chen2015multinomial,fildes2015effectiveness,li2014development}. In addition to the primary crash, secondary crashes have also been studied by researchers, who developed models based on the speed contour plot to predict the probability of secondary crashes \citep{xu2016real,park2016real}.

Although fruitful approaches have been proposed to improve the predictive performance, few studies focused on the issue of missing data in the real-time crash likelihood estimation. However, the approaches for missing data imputation utilized in other related areas, such as the missing traffic volume data estimation \citep{tang2015hybrid}, missing data imputation in road networks \citep{asif2013low}, may provide an insight for the missing data imputation in the traffic safety area. Traditional missing data imputing approaches utilized in the transportation area included the historical mean/median imputation, $k$-means clustering imputation, etc. \citep{conklin2003data,deb2016missing}. Recently, tensor decomposition has also been developed to impute missing data in various areas \citep{wu2017robust,tan2013tensor}.

The pattern of missing data means the distribution of missing values in the whole dataset. \citet{little2014statistical} classified missing patterns into three categories: missing completely at random (MCAR), missing at random (MAR), and not missing at random (NMAR).

\begin{enumerate}[(I)]
	\item MCAR means the missing data does not have any relationship with the distribution of the observations, while it does not depend on specific variables.
	\item MAR indicates that the distribution of missing data has no relationship with the missing values, but is related to the observed data of other attributes.
	\item NMAR refers to the cases that the distribution of missing data has a certain pattern. This is the most difficult mechanism to deal with because the missing pattern should be treated case by case.
\end{enumerate}

The MCAR or MAR problem could be addressed by some universal algorithms while the NMAR problem means almost countless possibilities in the distribution of missing values. Various kinds of algorithms, which assume that the missing pattern is MCAR or MAR, have been proposed for missing data imputation. These algorithms could mainly be divided into three categories \citep{li2004towards}:

\begin{enumerate}[(I)]
	\item The first group is discarding the samples with missing data, which is acceptable when the missing ratio is low and the data resource is abundant.
	\item The second group is interpolation, which refers to the process of interpolating missing values based on the existing information under certain regulations, such as, mean or median imputation, nearest imputation, and $k$-means imputation.
	\item The third group is EM (expectation maximization) based parameter estimation, which estimates the parameters of the data distribution through the existing data and then impute the missing data based on the estimated distribution.
\end{enumerate}

Among the methods in the third group, principal component analysis (PCA) based methods are considered as a category of efficient and reliable missing data imputation methods, which incorporate PCA and the EM estimation \citep{ilin2010practical}. It was proved that the probabilistic principal component analysis (PPCA) based missing data imputation method outperformed the conventional methods (e.g., the nearest/mean historical imputation methods and the local interpolation/regression methods) in the traffic flow volume dataset \citep{qu2009ppca}, mainly due to the two reasons: (I) the data of traffic flow volume followed the Gaussian distribution, which was in accordance with the hypothesis of the PCA-based missing data imputation; (II) PPCA succeeded at combining and utilizing global information as well as the local information.

\cite{dear1959principal} was the first to develop a PCA-based formulation to impute missing data. \cite{grung1998missing} further employed a least-square approach to solve the problem, and thus the algorithm was named the least-square PCA (LS-PCA). However, LS-PCA faced the overfitting issue, especially when the missing ratio was high. To solve this problem, \cite{tipping1999probabilistic} proposed the PPCA algorithm, which added a generalization term to the objective function to avoid overfitting, by assuming that the PCA followed a probabilistic form. Furthermore, the variational Bayesian PCA (VBPCA) was proposed to address the sensitive dependence on initial values of parameters, which was frequently observed in PPCA \citep{bishop1999variational}. \cite{ilin2010practical} summarized these three algorithms and compared their imputing performance in artificial experiments. In the domain of transportation, \citet{qu2009ppca,li2013comparison} developed a broad range of variants of PCA-based missing data imputation, including PPCA and kernel probabilistic principle component analysis (KPPCA), which demonstrated outstanding performance on resolving issues of missing values in traffic volume estimation.

PCA-based approaches have at least three merits in the domain of missing traffic data imputation. Firstly, it does not require strict assumptions such as the daily similarity, no continuous incompleteness of data points, and a large database. Secondly, the principal components remove the relatively trivial details and make sure that only the major information is used for constructing the probabilistic distribution of the latent variables. Thirdly, it simultaneously achieves the high imputing accuracy, acceptable speed, and robustness to abnormal data points in a broad range of missing traffic data imputation issues.

Although the cutting-edge PCA-based imputation algorithms have been utilized to impute the missing data in the traffic volume, there exist differences between the traffic flow data and the dataset in real-time crash likelihood estimation. Traffic flow data have continuous time-series properties, such as periodicity and tendency. On the contrary, the dataset of real-time crash likelihood estimation can be viewed as a table with rows of samples and columns of features. The values of the features in different rows are not in sequential orders and do not have periodicity and tendency, which indicates that imputing the missing data in real-time crash likelihood estimation is more difficult than that in the traffic flow data analysis. Most of the imputing approaches, such as the historical mean/median imputation, $k$-means clustering methods and interpolations, rely on the assumptions of periodicity and tendency, which no longer hold in the imputation of the crash data table. However, PCA-based imputation approaches do not have strict assumptions on the periodicity and continuity of the data; they first extract the major and able-to-model information and discard the trivial and unable-to-model details via principal components, and then use the obtained domain probabilistic distributions to impute missing values via MLE. Thus, PCA-based imputation approaches are assumed to have better performance in imputing missing crash data. One of the main goals of this paper is to verify this assumption and examine the PCA-based algorithms' ability in imputing missing data within real-time crash likelihood estimation.

Classification on the imbalanced dataset is another common issue in real-time crash likelihood estimation but has drawn little attention. In a binary classification problem, a dataset is said to be imbalanced when the number of samples in one class is higher than the other one \citep{seiffert2010rusboost,guo2008class}. The class with more samples is named the major class while the other with relatively fewer samples is denoted as the minor class. The imbalanced issue is commonly observed in a broad range of classification problems \citep{longadge2013class}, e.g., medical diagnosis detection of rare disease, determining frauds in banking system, detecting failures of technical devices, etc.

Real-time crash likelihood prediction is a typical imbalanced classification problem since the number of crash cases is usually much smaller than that of non-crash cases. This issue has attracted researchers' attentions in recent two years. \cite{theofilatos2016predicting} considered accidents as rare-events and developed a series of rare-event logit models to predict real-time accidents. \cite{basso2018real} proposed an accident prediction model combining SMOTE and SVM, which was then validated with original imbalanced data instead of artificially balanced data. To mitigate the imbalanced issue, \cite{yuan2017predicting} proposed an informative sampling approach that selected diverse negative (non-crash) samples, with some close to and some far from positive (crash) samples. In this paper, we use the matched case control method \citep{ahmed2013application} to select the negative samples. The previous research commonly selected 4:1 as the ratio of the number of non-cash cases to crash cases \citep{ahmed2012viability,ahmed2013data}. To better illustrate the performance of various solutions to imbalanced issues, we use a larger ratio 10:1 for demonstration purposes, which means that one crash case is matched with 10 non-crash cases.

In an imbalanced classification problem, most of the classifiers tend to classify all the samples into the major class, which sacrifices the accuracy of predicting a sample from the minor class. This is due to the inherit motivation of their objective functions which minimize the sum of errors by assigning the same weight to both the major and minor classes, where samples in the minor class make little contribution \citep{sun2007cost}. There are two main solutions to this problem: (I) Cost-sensitive learning techniques which stimulate the classifiers to pay more attention to the minor class by assigning different weights to the samples of both classes in the objective function \citep{pazzani1994reducing}; (II) Re-sampling the dataset, including over-sampling and under-sampling \citep{chawla2004editorial,estabrooks2000combination,kubat1997addressing}. SMOTE, viewed as one of the most efficient re-sampling algorithms, over-samples the minor class by generating synthetic samples instead of simple replications \citep{chawla2002smote}.

\section{Methodology}

As mentioned above, the main objectives of this paper are to seek out the most suitable imputation approaches and solutions to imbalanced issues in the domain of real-time crash risk estimation, and to examine different classifiers' tolerance to missing data. In this section, we first revisit the problem of real-time crash likelihood estimation and related classical classifiers (Problem 1), then present the two solutions to imbalanced issues (Problem 2), and finally propose the PCA-based missing data imputation approaches (Problem 3).

\subsection{Real-time crash likelihood estimation}

Most of the studies in the domain of real-time crash likelihood estimation define the problem as a binary classification problem.

\textbf{Problem 1} \emph{(Real-time crash likelihood estimation)}: Given a training dataset of $m$ samples, $({\bf{x}}^{(1)},y^{(1)}),...,({\bf{x}}^{(i)},y^{(i)}),...,({{\bf{x}}^{(m)}},y^{(m)}),i =1,...,m$, where ${\bf{x}}^{(i)}$ means the vector of explanatory variables of the $i^{th}$ sample and $y^{(i)} \in \{-1,1\}$, the object is to find a function $f({\bf{x}};  \boldsymbol{\omega}, b)$ which can distinguish positive $y^{(i)}$ from negative $y^{(i)}$. In this paper, the positive labels $y^{(i)}=1$ refers to crash cases, while the negative labels $y^{(i)}=-1$ refers to non-crash cases.

\subsubsection{Classical classifiers}

A broad range of machine learning algorithms have been used to detect crash-prone cases. However, the main focus of this paper is to incorporate missing data imputation and solutions to imbalanced issues in real-time crash likelihood estimation, instead of developing new classifiers. In this paper, we only consider two classical classifiers: SVM (with linear, Gaussian, and polynomial kernels) and AdaBoost ensemble algorithm.

Firstly, SVM constructs a hyperplane or a set of hyperplanes in high-dimensional space to separate data samples into two kinds of labels. Intuitively, a good hyperplane aims to maximize the distance to the nearest data points of binary classes, while it also extends to the training dataset that cannot be linearly separated by introducing slack variables. Starting from this idea, the general objective function of SVM can be written as Eq. (\ref{eq:1}). The first part of the equation refers to an L2 norm regularization, while the second part is in a form of hinge loss.

\begin{equation} \label{eq:1}
\mathop {{\rm{min}}}\limits_{{\boldsymbol{\omega}} ,b} \;\;\frac{1}{2}{ || {\boldsymbol{\omega}}|| ^2} + C \sum\limits_{i=1}^m {{\rm{max}}(0,1-{y^{(i)}}(\boldsymbol{\omega}^{\textrm{T}}\boldsymbol{x}^{(i)}+b))}
\end{equation}

In order to apply SVM to a non-linear separable classification problem, the kernel trick can be utilized. In this paper, three categories of kernels are used:

\begin{enumerate}[(I)]
	\item Linear kernel: $K({\bf{x}}^{(i)}, {\bf{x}}^{(j)}) = \left\langle {{{\bf{x}}^{\left( i \right)}},{{\bf{x}}^{\left( j \right)}}} \right\rangle$;
    \item Gaussian kernel: $ K({\bf{x}}^{(i)}, {\bf{x}}^{(j)}) = \exp (-|| {\bf{x}}^{(i)} - {\bf{x}}^{(j)} ||^2)$;
    \item Polynomial kernel: $ K({\bf{x}}^{(i)}, {\bf{x}}^{(j)}) = (1+\left\langle {{{\bf{x}}^{\left( i \right)}},{{\bf{x}}^{\left( j \right)}}} \right\rangle)^p$, specifically, $p=3$ is selected in this paper. \\
\end{enumerate}

Secondly, boosting is a combination of machine learning techniques which ensemble a series of weak learners, each of which achieves a classification accuracy slightly larger than 0.5 for the binary classification, to improve the prediction performance and robustness. AdaBoost, the abbreviation of ``Adaptive Boosting'', is a typical boosting method which adapts the subsequent weak learners to emphasizing more on the samples misclassified by the prior weak learners. AdaBoost generates $T$ weak learners $h_1 ({\bf{x}}^{(i)}),..., h_t({\bf{x}}^{(i)}),...,h_T ({\bf{x}}^{(i)})$ and assembles them into a strong learner by multiplying each weak learner with a coefficient $\alpha_t$, see Eq. (\ref{eq:3}).

\begin{equation} \label{eq:3}
{F_T} ( {\bf{x}}^{(i)}) = \mathop \sum \limits_{t = 1}^T {\alpha _t}{h_t} ( {\bf{x}}^{(i)}), \ \ \ t = 1,\ldots ,T
\end{equation}
where $ h_t ({\bf{x}}^{(i)}): {\bf{x}}^{(i)} \mapsto \{-1,+1\}$. $h_t$ can be any classifier, such as the decision tree, LR, SVM and so on.

\subsubsection{Measures of Effectiveness (MoEs)}

Accuracy is the basic and simplest MoE in the classification problem, but it may produce a biased illusion on imbalanced data. In the case that the ratio of the number of negative samples to the number of positive samples is large, e.g., 99:1, the classifier may predict all the samples into the negative class, consequently, the accuracy achieves 99\%. However, this result is meaningless to scenarios where the objective is to detect the minority cases, such as the crash likelihood prediction. To overcome this biased phenomenon, the confusion matrix is utilized, where samples can be categorized into four conditions, i.e., true negative (TN), false negative (FN), false positive (FP), and true positive (TP), as shown in Table \ref{table:confusion}.

\begin{table}[t]
	\caption{The Confusion Matrix}
	\label{table:confusion}
	\centering
	\begin{tabular}{cccc}
		\toprule
		\toprule
		& & \multicolumn{2}{c}{True label} \\
		\cmidrule{3-4}
		& & 0 (non-crash) & 1 (crash) \\
		\midrule
		\multirow{2}{*}{\centering Predictive label} & 0 (non-crash) & TN & FN \\
		\cmidrule{2-4}
		& 1 (crash) & FP & TP \\
		\bottomrule
		\bottomrule
	\end{tabular}
\end{table}

Based on the confusion matrix, the true positive rate and false positive rate can be calculated by
\begin{equation}
{\rm{True \ Positive \ Rate: TPR}} = \frac{{{\rm{TP}}}}{{{\rm{TP + FN}}}}
\end{equation}
\begin{equation}
{\rm{False \ Positive \ Rate: FPR}} = \frac{{{\rm{FP}}}}{{{\rm{TN + FP}}}}
\end{equation}

Most of the classifiers provide probabilistic degrees or predicting scores of each sample; the higher the probabilistic degrees or predicting scores, the greater confidential level that the sample can be classified as a positive sample. By changing the score threshold, a group of confusion matrices and the corresponding TPR and FPR can be calculated. A receiver operation characteristic curve (ROC) shows the relationship between FPR on the X-axis and TPR on the Y-axis. The ROC curve considers both the accuracies of classifying positive samples and negative samples, and thus provides a more fair MoE for imbalanced data. To quantify the performance of the ROC curve, a higher AUC value (areas under ROC curve) implies a stronger classifier.

In addition to the AUC value, sensitivity and specificity are also utilized in this paper. Sensitivity is the same as TPR while specificity equals to (1-FPR). The high sensitivity indicates that the classifier successfully predicts the majority of the crash occurrence, while high specificity implies that the classifier produces less irrelevant alarms. In an imbalanced dataset, improving sensitivity usually means sacrificing specificity to some extent. Therefore, a trade-off should be made between these two MoEs.

\subsection{Solutions to Imbalanced Data Classification}

\textbf{Problem 2} \emph{(Solutions to imbalanced issue)}: Given an imbalanced dataset with much more samples from the majority class than those from the minority class, how to simultaneously achieve acceptable sensitivity, specificity, and AUC in the classification?

\subsubsection{Solution I---Algorithmic-Level Solutions}
On the algorithmic level, cost-sensitive learning is the most widely used method to take the misclassified cost of different classes into consideration. The cost-sensitive learning assigns different costs of misclassifying different classes. The objective functions of most classifiers can be written as the sum of two parts: one is the sum of empirical errors and the other is a regularization term, e.g., Eqs. (\ref{eq:1}--\ref{eq:3}). Although a coefficient $C$ is employed to measure the trade-off of the two terms, it is set to be the same for all the samples without distinguishing positive and negative labels, which may produce biased results towards the majority class. The cost-sensitive learning technique (COST) simply adjusts this biased classification by allocating a larger coefficient to the minority class, which means misclassifying a sample from the minority class (crash case in this paper) receives a higher penalty. For instance, the objective function of SVM in Eq. (\ref{eq:1}) can be rewritten as follows:

\begin{multline} \label{eq:imbalanced}
\mathop {{\rm{min}}}\limits_{{\boldsymbol{\omega}} ,b} \;\;\frac{1}{2}{ || {\boldsymbol{\omega}}|| ^2} + {C_{\textrm{}{pos}}} \sum\limits_{i=1,i \in Pos}^m {{\rm{max}}(0,1-{y^{(i)}}(\boldsymbol{\omega}^{\textrm{T}}\boldsymbol{x}^{(i)}+b))} + \\
 {C_{\textrm{}{neg}}} \sum\limits_{i=1,i \in Neg}^m {{\rm{max}}(0,1-{y^{(i)}}(\boldsymbol{\omega}^{\textrm{T}}\boldsymbol{x}^{(i)}+b))}
\end{multline}
where $Pos$ and $Neg$ are the sets of positive and negative training samples, respectively, and $C_{pos}$ and $C_{neg}$ are the corresponding coefficients. This transformation can be easily migrated to other classifiers like LR, decision tree, etc. In the following section, a sensitivity analysis is conducted to examine MoEs of the classifiers under different ratios of $C_{pos}$ to $C_{neg}$.

\subsubsection{Solution II---Data-Level Solutions}

To avoid the poor performance of classifiers on imbalanced data, re-sampling at the data level is another group of approaches, which can be classified into two types: over-sampling and under-sampling. Simple over-sampling may lead to overfitting issues while under-sampling will discard a large amount of potentially meaningful information in the small dataset. To overcome these drawbacks, an advanced over-sampling algorithm called the synthetic minority over-sampling technique was proposed by \cite{chawla2004editorial}. SMOTE creates ``synthetic'' samples of the minority class instead of simply duplicating the existing samples.

\subsection{PCA-based missing data imputation}

\textbf{Problem 3} \emph{(Imputing missing crash data table)}: Given an incomplete crash table with rows of samples and columns of features, how to impute the missed values based on the observed values?

The PCA-based missing data imputation algorithms formulate the relationship between original variables and latent variables in a PCA-based form, and then solve the problem with EM iterations. PCA is a machine learning technique which can compress high-dimensional data into low-dimensional data with the minimum loss on variance. This low-dimensional data after compression can also be reconstructed into the original data. This property can be utilized in missing data imputation: first estimate the probability distribution of the compressed information based on the original observed data, and then reconstruct the missing data by the compressed information, which can also be viewed as latent variables. PCA-based missing data imputation consists of three main kinds of algorithms, including LS-PCA, PPCA, and VBPCA, which have different assumptions on the relationships between original variables and latent variables.

Suppose that we have $m$ samples of $d \times 1$ original vectors ${\bf{t}}_1,{\bf{t}}_2,...,{\bf{t}}_m$, which can be formulated as a function of $c \times 1$ dimensional latent variables:
\begin{equation}
{{\bf{t}}_j} = {\bf{W}}{{\bf{z}}_j} + {\boldsymbol{\mu}}
\end{equation}
where ${\bf{W}}$ is a $d \times c$ matrix, ${\bf{z}}_j$ is a $c \times 1$ vector of principal components (i.e. latent variables), and ${\boldsymbol{\mu}} $ is a $d \times 1$ bias term.

\subsubsection{Imputation Algorithm I---Least-Square PCA (LS-PCA)}
A straightforward method to determine the latent variables is to minimize the mean-square error between the reconstructed ${\hat{t}}_{ij}$ attained from latent variables and the original observed $t_{ij}$:

\begin{equation} \label{eq:LSPCA1}
\min \;\;\mathop \sum \limits_{i,j \in O} {\left( {{t_{ij}} - {{\hat t}_{ij}}} \right)^2}
\end{equation}
\begin{equation} \label{eq:LSPCA2}
{{\hat t}_{ij}} = {\bf{w}}_i^{\rm{T}}{{\bf{z}}_j} + {\mu _i} = \mathop \sum \limits_{k = 1}^c {w_{ik}}{z_{kj}} + {\mu _i}
\end{equation}
where $t_{ij}$ means the $i$th variable of the $j$th sample of the observed data, while ${\hat{t}}_{ij}$ is the reconstruction of the data element $t_{ij}$. $O$ is the set of indexes $i,j$. $z_{kj}$ means the $k$th latent variable of the $j$th sample of the latent space.

This optimization problem can be solved by a least-square algorithm which updates parameters $\bf{W}$, $\boldsymbol{\mu}$, and latent variables ${\bf{z}}_j$. However, the LS-PCA algorithm might easily suffer from the overfitting issue, especially when the missing ratio is high, since the object of LS-PCA is to minimize the mean square error between the observed original data and the reconstructed data, thus the algorithm may generate unreasonable large parameters to well fit observed data and lose the generalization ability.

\subsubsection{Imputation Algorithm II---Probabilistic PCA (PPCA)}
A natural solution to the overfitting problem of LS-PCA is adding a regularization term in the objective function to penalize unreasonably large parameters. Another solution is altering the transformation between the original data and latent variables to a probabilistic form, from which the regularization term is naturally derived. PPCA is derived by adding an isotropic term to Eq. (\ref{eq:LSPCA1}):
\begin{equation}
{{\bf{t}}_j} = {\bf{W}}{{\bf{z}}_j} + {\boldsymbol{\mu}} + {\boldsymbol{\varepsilon}} _j
\end{equation}
where ${{\bf{z}}_j}$ and ${\boldsymbol{\varepsilon}} _j$ follow the normal distributions, i.e., ${{\bf{z}}_j} \sim \mathcal{N}({\bf{0}},{\bf{I}})$, ${\boldsymbol{\varepsilon}} _j \sim \mathcal{N}({\bf{0}},v{\bf{I}})$. There are three groups of parameters, i.e., ${\bf{W}}$, ${\boldsymbol{\mu}}$ and $v$, which can be estimated by the EM algorithm \citep{bishop1999variational}.

\subsubsection{Imputation Algorithm III---Variational Bayesian PCA (VBPCA)}
PPCA is sometimes sensitive to the initialization of parameters ${\bf{W}}$, ${\boldsymbol{\mu}}$ and $v$. To overcome this defect, an assumption of the Gaussian prior probabilistic distribution was made to the parameters ${\bf{W}}$ and ${\boldsymbol{\mu}}$, which formulates the VBPCA. ${\bf{W}}$ and ${\boldsymbol{\mu}}$ follow the normal distributions: ${\boldsymbol{\mu}} \sim \mathcal{N}({\bf{0}},v_{\mu}{\bf{I}})$, ${\bf{w}}_i \sim \mathcal{N} ({\bf{0}}, v_{w,k} {\bf{I}})$, where $v_m$ and $v_{w,k} $ are the hyperparameters that can be updated during learning (e.g., using the evidence framework or variational approximations). VBPCA can also be iteratively solved by the EM algorithm \citep{ilin2010practical}.

To summarize, in the PCA framework, LS-PCA employs the least-square approach to estimate the parameters, which might be overfitting when the missing ratio of traffic flow data is high. PPCA introduces a probabilistic form to the original data by adding an isotropic term to penalize the occurrence of unreasonable large parameters, and thus avoids overfitting issues. VBPCA further introduces a probabilistic form to the parameters in order to eliminate the variance of the initialization of the parameters in PPCA.

\section{Numerical Tests and Discussions}

\subsection{Data Preparation}

\begin{figure}[!t]
	\centering
	\subfloat[Layout of study site]{\includegraphics[width=\linewidth]{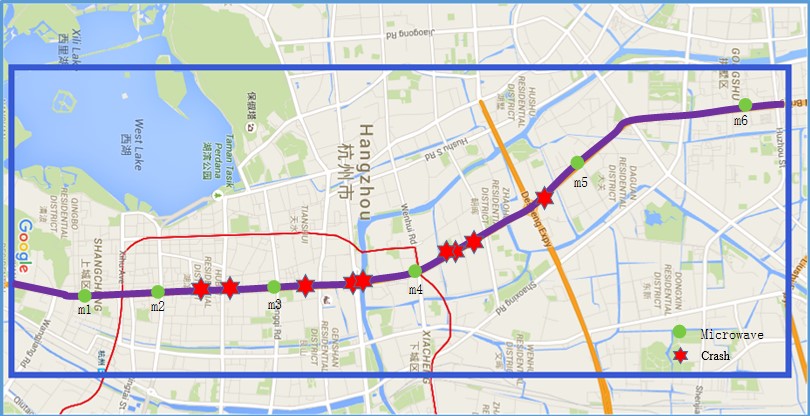}%
		\label{fig1a}}
	\\
	\subfloat[Sensor and crash locations]{\includegraphics[width=0.8\linewidth]{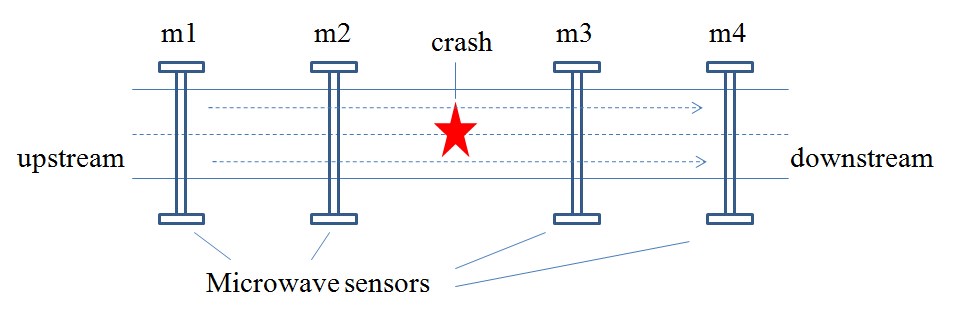}%
		\label{fig1b}}
	\\
	\subfloat[Time intervals]{\includegraphics[width=0.8\linewidth]{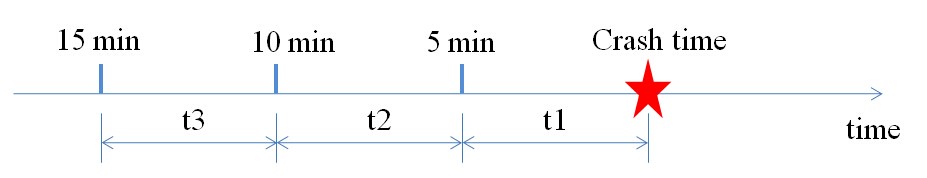}%
		\label{fig1c}}
	\caption{Microwave sensor locations and crashes of the study site.}
	\label{fig1}
\end{figure}

The data used in this study were obtained from the Traffic Management Platform of Hangzhou, China. This platform continuously collects 5-min aggregated lane-by-lane traffic flow, occupancy, and speed via remote traffic microwave sensors for each lane in real time. Besides the microwave sensor data, the platform also provides traffic crash records with several attributes including the incident time, location, and incident type. The study site is located at the Shangtang-Zhonghe Urban Expressway of Hangzhou (see Fig. \ref{fig1a}). Two datasets (a train-and-test dataset and a validation dataset) are collected and each dataset has a crash record table and traffic flow database collected from the microwave sensor data.

Each crash case in the table of crash records is mapped with 2 closest upstream and 2 closest downstream microwave sensors. For each crash case, we collect the dynamic traffic flow data in 5-10 min and 10-15 min prior to its occurrence time, respectively.

Fig. \ref{fig1b} shows the 4 microwave sensors selected for each crash and their names: m1 means the second closest upstream sensors to the crash, m2 means the nearest upstream sensors to the crash, and so on. Fig. \ref{fig1c}. shows the time intervals prior to each crash, among which t2 (5-10 min prior to the crash occurrence) and t3 (10-15 min prior to the crash occurrence) are collected in our study because t1 does not provide enough time for alarm or other crash-prevention measures based on the crash likelihood prediction.

\begin{table}[!t]
	\caption{Number and Variable Name of 24 Explanatory Variables}
	\centering
	\label{variables}
	\begin{tabularx}{0.95\textwidth}{ccccccc}
		\toprule
		\toprule
		Number & 1 & 2 & 3 & 4 & 5 & 6 \\
		Variable & f-m1-t3 & f-m1-t2 & f-m2-t3 & f-m2-t2 & f-m3-t3 & f-m3-t2 \\
		\midrule
		Number & 7 & 8 & 9 & 10 & 11 & 12 \\
		Variable & f-m4-t3 & f-m4-t2 & o-m1-t3 & o-m1-t2 & o-m2-t3 & o-m2-t2 \\
		\midrule
		Number & 13 & 14 & 15 & 16 & 17 & 18 \\
		Variable & o-m3-t3 & o-m3-t2 & o-m4-t3 & o-m4-t2 & s-m1-t3 & s-m1-t2 \\
		\midrule
		Number & 19 & 20 & 21 & 22 & 23 & 24 \\
		Variable & s-m2-t3 & s-m2-t2 & s-m3-t3 & s-m3-t2 & s-m4-t3 & s-m4-t2 \\
		\bottomrule
		\bottomrule
	\end{tabularx}
\end{table}

Three kinds of variables, i.e., flow, time occupancy and speed, are selected for each crash case, thus each sample contains 4 (sensors) $\times$ 2 (time intervals) $\times$ 3 (variable types) = 24 variables. The notation of the explanatory variables is shown in Table \ref{variables}, where the first part of each variable indicates the category (e.g., flow, occupancy and speed), the second part refers to the RTMS, and third part represents the time interval. For example, f-m1-t3 represents the flow (f) at the second closest upstream sensor (m1) during the time interval of 5-10 min prior to the crash (t2).

As above-mentioned, the matched case-control strategy is applied in selecting the non-crash samples. In this paper, we match 10 non-crash samples for each crash sample, where the matching rules are illustrated as follows:

\begin{enumerate}[(I)]
	\item \textit{Location}. The location of the matched non-crash cases should be the same as the crash case;
	\item \textit{Within-day time}. The within-day time of the non-crash cases should be the same as that of the matched crash case, but they should be in different days. For example, if one crash occurred at 12:45 PM on June 15, 2015 (Monday), then one matched non-crash case can be extracted from the dataset with the time stamp of 12:45 PM on June 29, 2015 (Monday);
	\item \textit{Day type}. We define two kinds of day types, i.e., weekday and weekend, and the crash cases should share the same day type with the matched non-crash case.
\end{enumerate}

The train-and-test dataset collected from June 11 to November 11, 2015, is used for the training and testing (10-fold cross validation in this paper). In order to implement the sensitivity analysis of the missing data, we only select the crash cases which can be matched with complete explanatory variables in the train-and-test dataset. The train-and-test dataset is iteratively split into 90\% training set and 10\% testing set in a 10-fold cross validation. On the other hand, the validation dataset is collected from June 1 to October 1, 2016, and used for validating the real effectiveness of the proposed framework, which is comprised of PCA-based missing data imputation and solutions to the imbalanced data (see Table \ref{table:dataset}). All the crash records in the validation dataset have been selected and matched with explanatory variables, while the missing ratio of the validation dataset reaches 21\%. The values of each explanatory variable in the two datasets are standardized, with a mean value of 0 and standard deviation of 1.

\begin{table}[!t]
	\caption{The train-and-test dataset and validation dataset}
	\label{table:dataset}
	\centering
	\small
	\begin{tabularx}{1\textwidth}{ccc}
		\toprule
		\toprule
		dataset & train-and-test dataset & validation dataset \\
		\midrule
		time range & 2015/06/11-2015/11/11 & 2016/06/01-2016/10/01 \\
		number of crash samples & 123 & 120 \\
		number of non-crash samples & 1230 & 1200 \\
		proportion of missing data & 0\% & 21\% \\
		\bottomrule
		\bottomrule
	\end{tabularx}
\end{table}

\begin{figure}[!ht]
	\centering
	\includegraphics[width=0.9\linewidth,trim=4 4 4 4,clip]{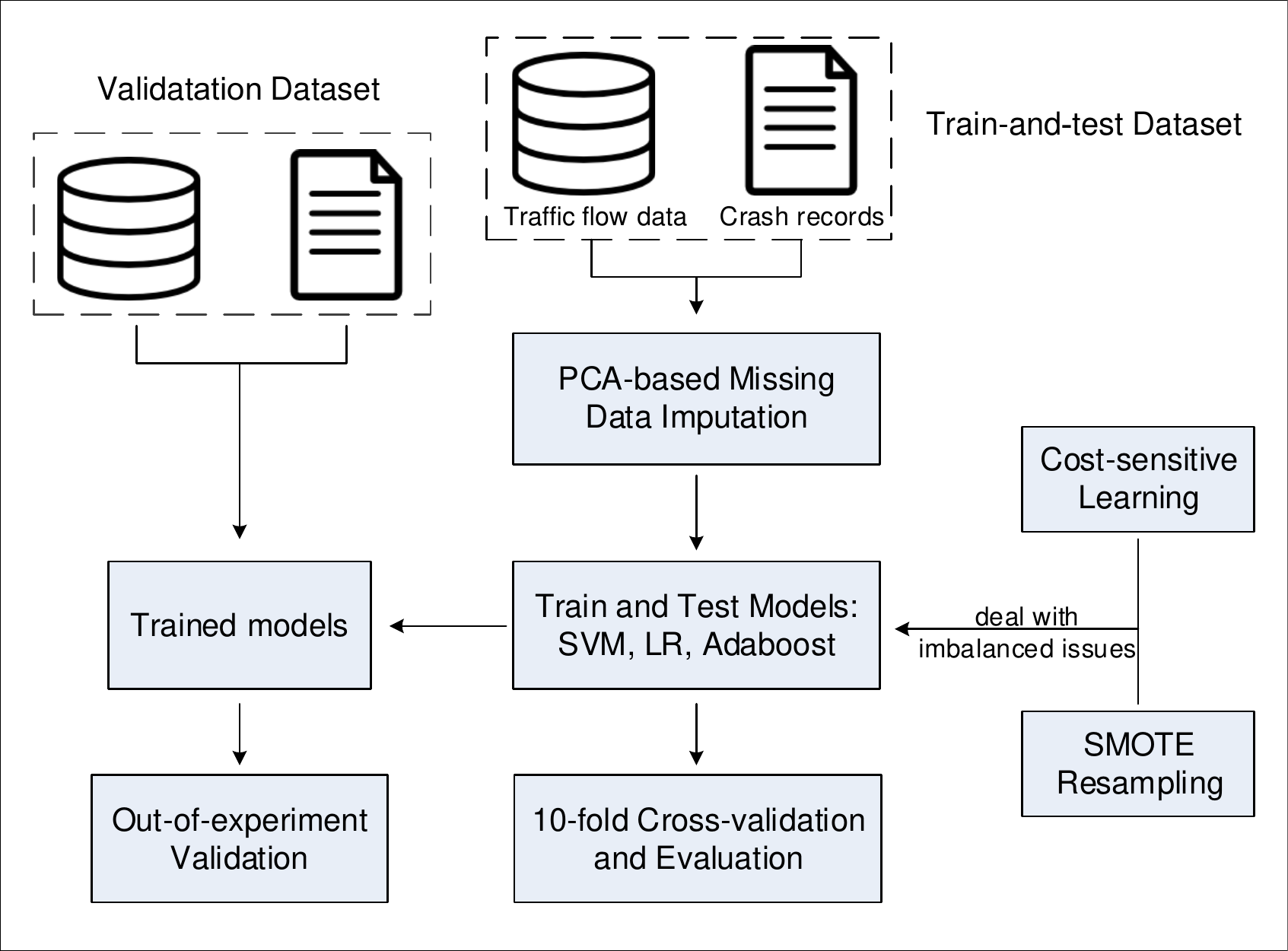}
	\caption{The framework.}
	\label{fig2}
\end{figure}

Fig. \ref{fig2} depicts the proposed framework which is comprised of three important parts: the predictive models for real-time crash likelihood estimation, the algorithms for missing data imputation, and two solutions to imbalanced issues. A series of sensitivity analyses are implemented to examine the efficiency and interaction of these three components based on the train-and-test dataset, and the optimal method for each component is selected. In addition, an out-of-experiment validation test is conducted to evaluate the predictive performance of the selected framework, based on the independent validation dataset.

\subsection{Sensitive Analysis of Solutions to Imbalanced Issues}

In this section, we design a sensitivity analysis to investigate how different class-weighted ratios affect the model MoEs in the train-and-test dataset, in terms of different solutions to imbalanced issues, including COST, SMOTE sampling, and the combination of them. Considering the real imbalanced ratio in the dataset is 1:10, five class-weighted ratios (defined as $\gamma$), i.e. 1, 5, 10, 20, 30, are tested in this sensitivity analysis. The parameters in solutions to imbalanced issues are depicted in Table \ref{table:imblanced}.

\begin{figure}[!ht]
	\centering
	\subfloat[Accuracy]{\includegraphics[width=0.5\textwidth]{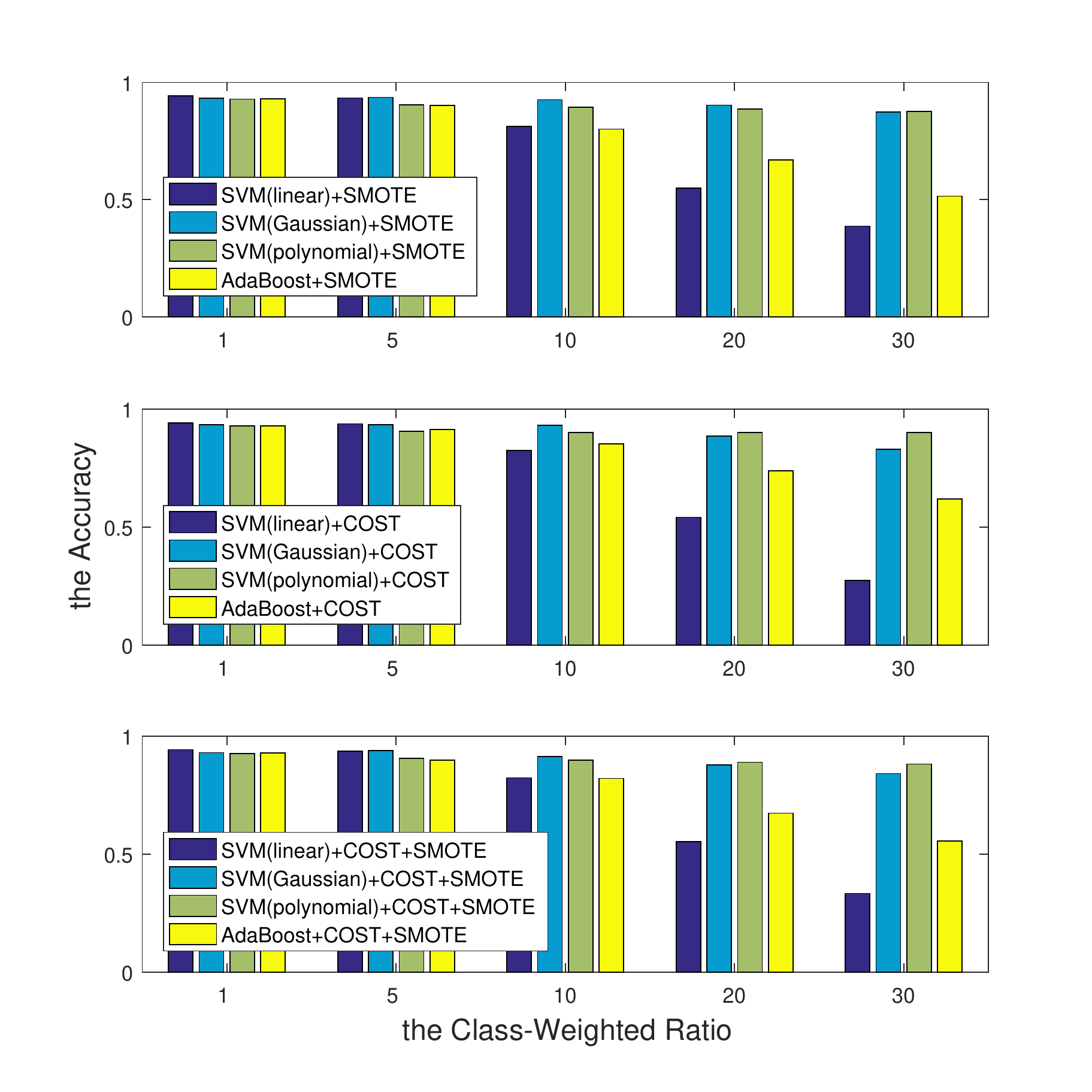}%
		\label{fig3a}}
	\hfil
	\subfloat[AUC]{\includegraphics[width=0.5\textwidth]{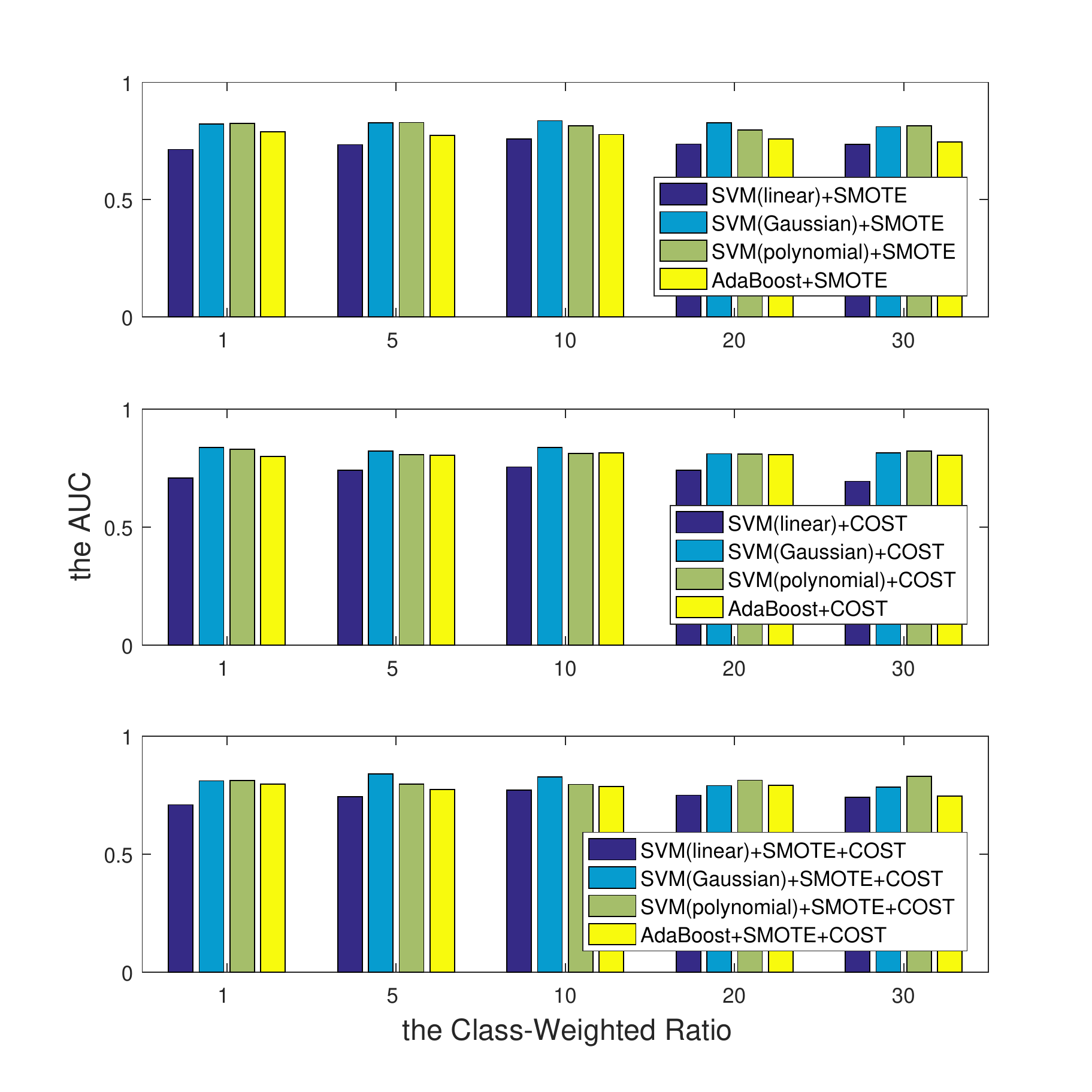}%
		\label{fig3b}}
	\\
	\subfloat[Sensitivity]{\includegraphics[width=0.5\textwidth]{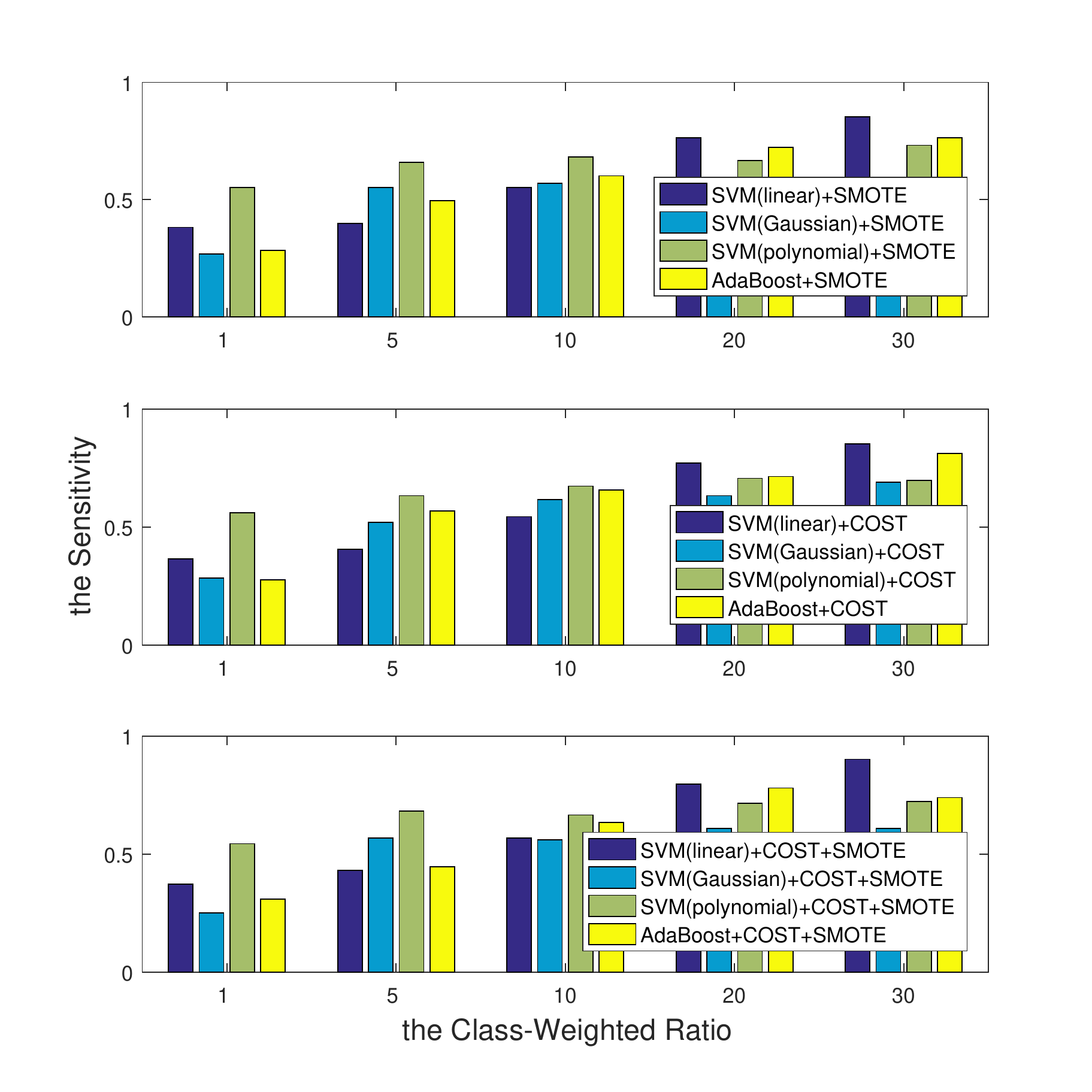}%
		\label{fig3c}}
	\hfil
	\subfloat[Specificity]{\includegraphics[width=0.5\textwidth]{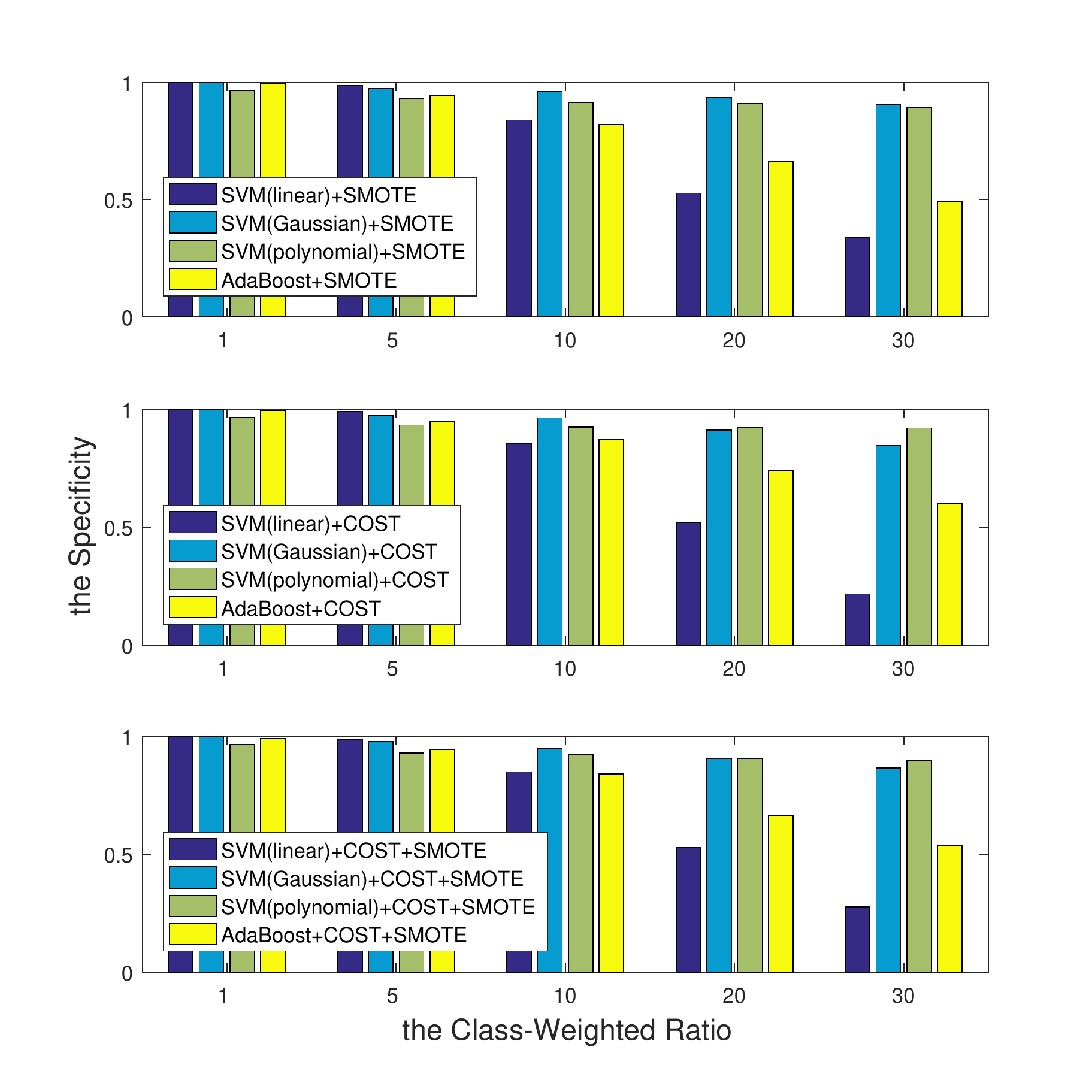}%
		\label{fig3d}}
	\caption{Sensitivity analysis on the class-weighted ratios}
	\label{fig3}
\end{figure}

Fig. \ref{fig3}. shows the accuracy, AUC, sensitivity and specificity of the 4 classifiers, i.e., SVM (with linear, Gaussian, and polynomial kernels), and AdaBoost, under different class-weighted ratios and different solutions to imbalanced issues. Some interesting results are observed:

\begin{table}[!t]
	\caption{Parameters in solutions to imbalanced issues}
	\label{table:imblanced}
	\centering
	\small
	\begin{tabularx}{0.9\textwidth}{cc}
		\toprule
		\toprule
		Solutions & Parameters \\
		\midrule
		COST &  $ C_{pos} / C_{neg} = \gamma $\\
		SMOTE &  Multiplier of synthetic samples $= \gamma$ \\
		COST + SMOTE & \begin{tabular}{@{}c@{}} In COST: $C_{pos} / C_{neg} = \sqrt{\gamma} $ \\In SMOTE:  the multiplier of synthetic samples $= \sqrt{\gamma} $ \end{tabular}\\
		\bottomrule
		\bottomrule
	\end{tabularx}
\end{table}

\begin{enumerate}[(I)]
	\item Accuracy of all models drops with the increase of the class-weighted ratio, which means the classifiers scarify a part of the precision and overall accuracy to improve the sensitivity; A trade-off should be made between the specificity and sensitivity because a low specificity means more false alarms, which will relax people's vigilance, while a low sensitivity indicates the low accuracy on identifying a real crash case;
	\item AUC is not sensitive to the imbalanced data because of its inherit property, and thus it is not a suitable indicator for selecting a proper class-weighted ratio;
	\item Linear classifier, i.e. SVM (linear), is more sensitive to the class-weighted ratio compared to non-linear classifiers, such as SVM (Gaussian), SVM (polynomial) and AdaBoost;
	\item Three examined solutions to imbalanced issues, i.e., COST, SMOTE, and COST + SMOTE, show comparable abilities in changing the classifiers' behavior under the same class-weighted ratio $\gamma$.
\end{enumerate}

In the following analyses, the class-weighted ratio is set to be 10 for all classifiers, by considering the trade-off between the sensitivity and specificity under such an imbalanced dataset.

\subsection{Missing Data Imputation: Accuracy Analysis}

In this section, we first discuss the trade-off between the computation complexity and imputation accuracy of the 3 PCA-based missing data imputing algorithms, and then compare them with two conventional interpolation methods, i.e., mean imputation, and $k$-means.

Two experiments are designed based on the train-and-test dataset, which is a complete dataset. The missing pattern is assumed to be MCAR. In Experiment I, 20\%, 40\%, and 60\% of the explanatory variables in the train-and-test dataset are randomly removed from each sample (for both crash cases and non-crash cases). It is noteworthy that different samples (rows) have different missing explanatory variables (columns). Each row and each column have at least one observed value, otherwise, the row or the column will be removed. Then three PCA-based missing data imputation approaches are utilized to impute the missing values. The dimensionality of latent variables is a key parameter in PCA-based approaches, thus the root mean squared error (RMSE) and computing time (which measures the computation complexity) are calculated under different latent dimensionality and different missing values. RMSE is calculated by

\begin{equation}
{\rm{RMSE}} = \sqrt {\frac{1}{N}\mathop \sum \limits_{i = 1}^N {{\left( {t_{real}^i - t_{impu}^i} \right)}^2}}
\end{equation}
where $t_{real}^i$ and $t_{impu}^i$ are the real and estimated values of the $i$th imputed values, respectively, while $N$ implies the number of imputed values.

After determining the latent dimensionality by Experiment I, the PCA-based approaches are compared with the two traditional interpolation imputing methods, i.e., mean imputation and $k$-means clustering imputation in Experiment II. Experiment II randomly generates a more elaborate group of missing ratios, which starts from 0 to 60\%, with a step of 5\%, in the train-and-test dataset. Considering that random generation of missing values leads to different results in different trails, both the Experiment I and Experiment II are repeated by 5 times and the mean results are presented, respectively,.

\subsubsection{Experiment I---Trade-off Between Computation Complexity and Imputing Accuracy}

\begin{figure}[!t]
	\centering
	\subfloat[PPCA]{\includegraphics[width=0.57\linewidth]{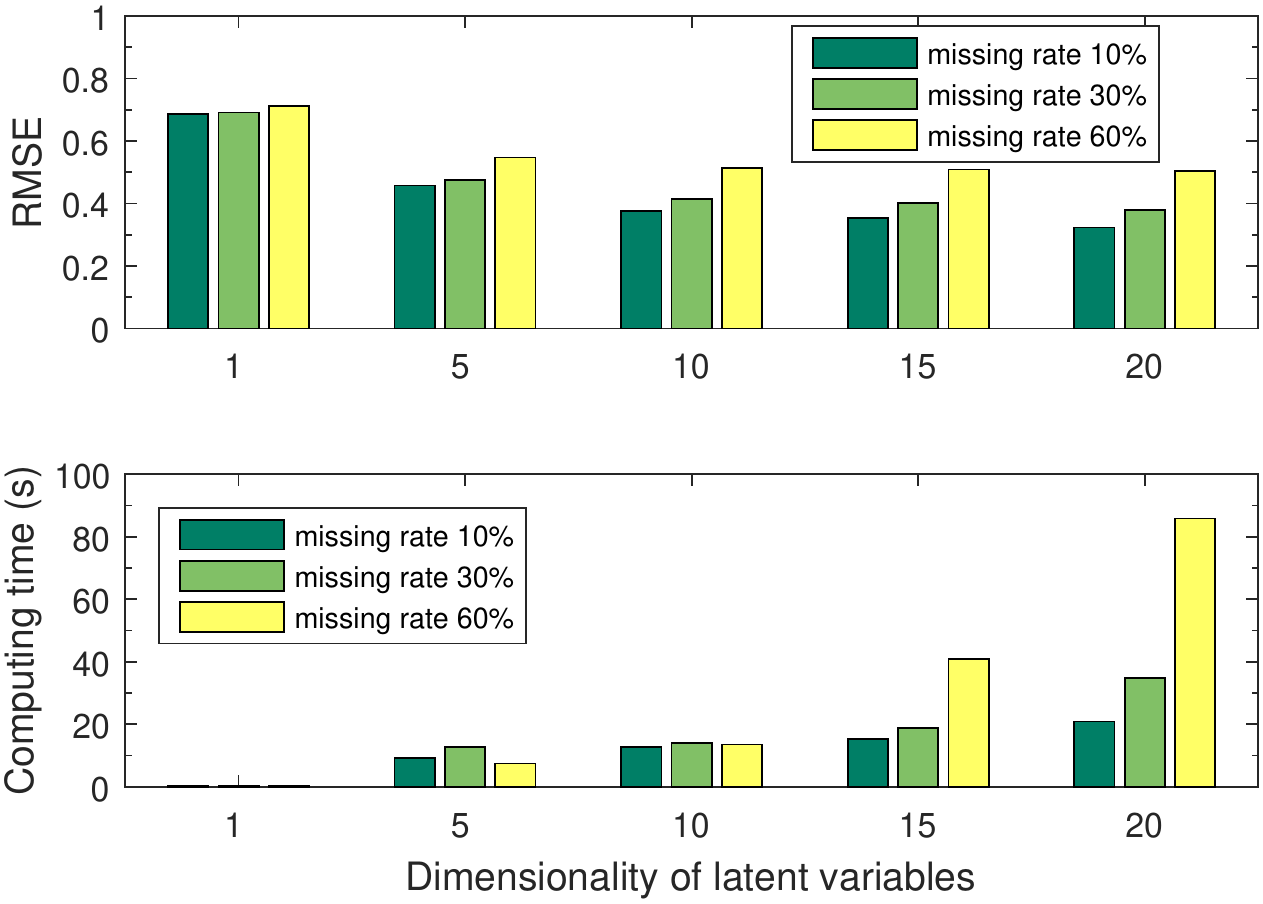}%
		\label{fig4a}}
	\\
	\subfloat[VBPCA]{\includegraphics[width=0.57\linewidth]{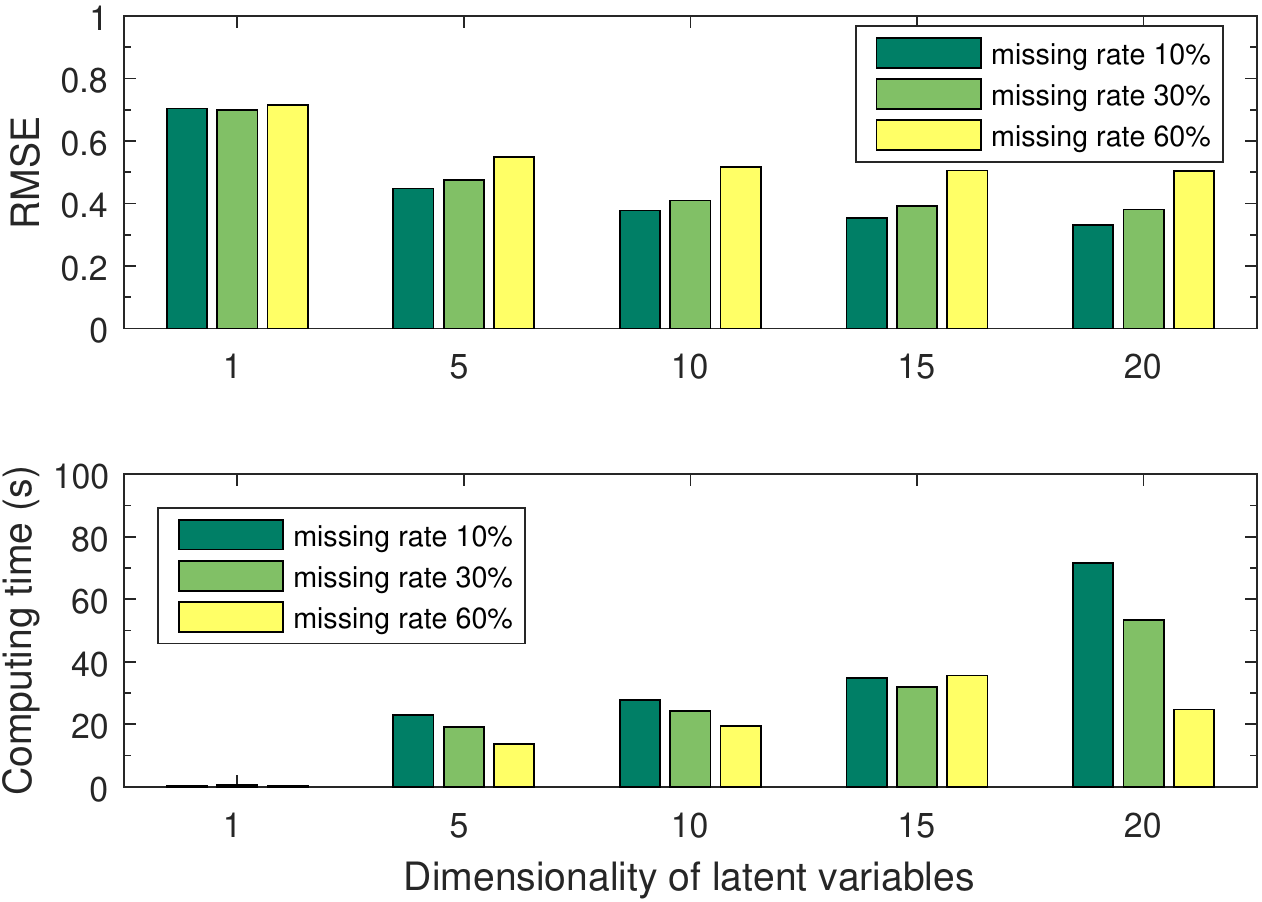}%
		\label{fig4b}}
	\\
	\subfloat[LS-PCA]{\includegraphics[width=0.57\linewidth]{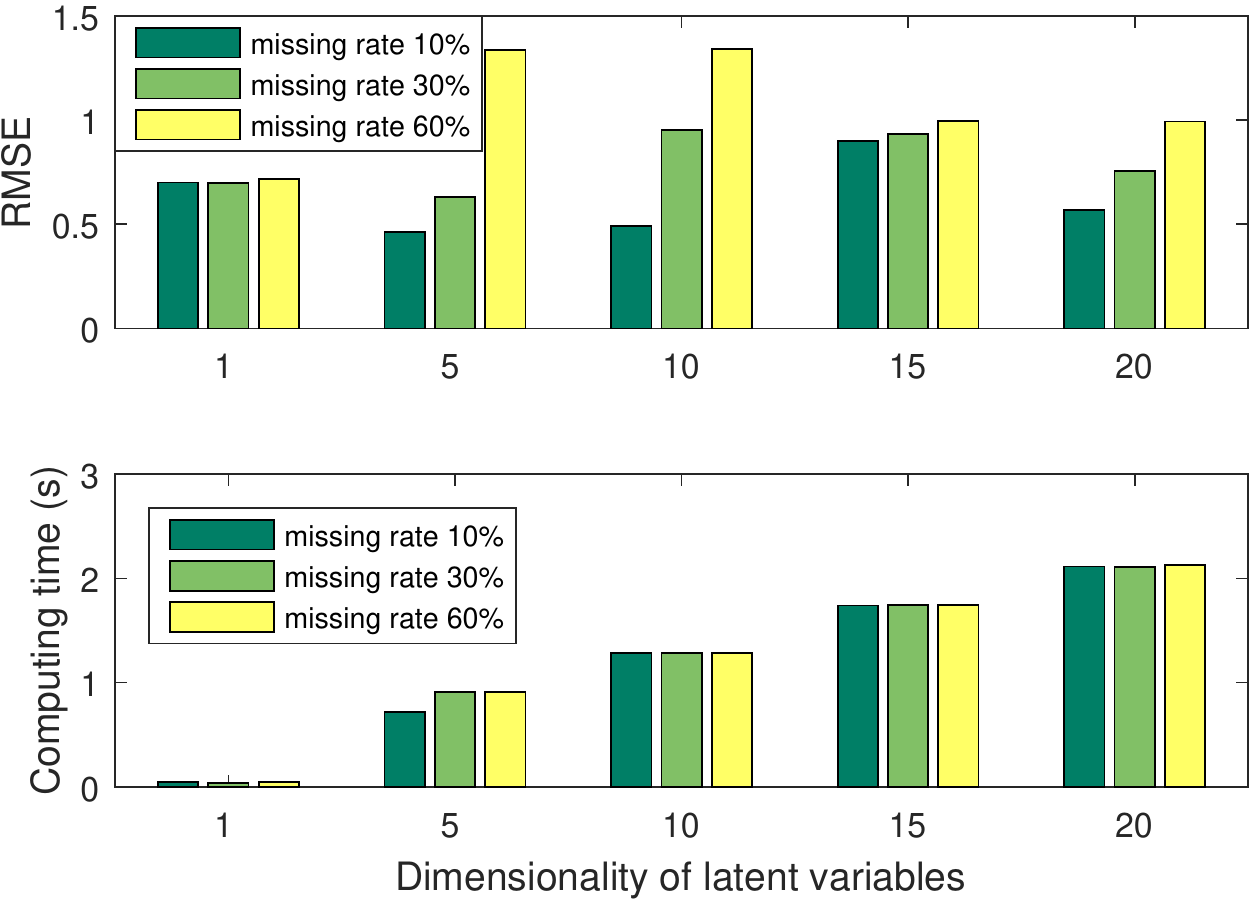}%
		\label{fig4c}}
	\caption{RMSE and computing time of missing data imputation algorithms.}
	\label{fig4}
\end{figure}

The results of \textit{Experiment I} are shown in Fig. \ref{fig4}. (I) RMSE decreases while the computing time increases with the dimensionality of latent variables, which indicates trade-off should be made between the accuracy and computation complexity. (II) PPCA and VBPCA achieve comparable imputing performance measured by RMSE and remain stable in different missing rates. (III) RMSE of LS-PCA shows a high fluctuation, especially in a high missing rate under which the RMSE can reach 1. This is caused by the the inherit overfitting issue of LS-PCA, whose objective function aims at minimizing the difference between the original data and reconstructed data without considering any regularization. It leads the algorithm to generate unreasonable large parameters.

By considering the trade-off between the computation complexity and imputation accuracy, 15 is selected as the latent dimensionality for the PCA-based imputing algorithms in this paper.

\subsubsection{Experiment II---Comparison Between PCA-based Imputation and Interpolations}

PCA-based missing data imputation algorithms belong to the EM based parameter estimation approach, where parameters are estimated based on observed data and then the missing data are imputed via the probabilistic distribution of these parameters. On the contrary, interpolation methods, e.g., mean imputation and $k$-means clustering imputation, aim to impute the missing values by the existing values based on specific rules of similarity.

\begin{figure}[!ht]
	\centering
	\includegraphics[width=0.5\linewidth]{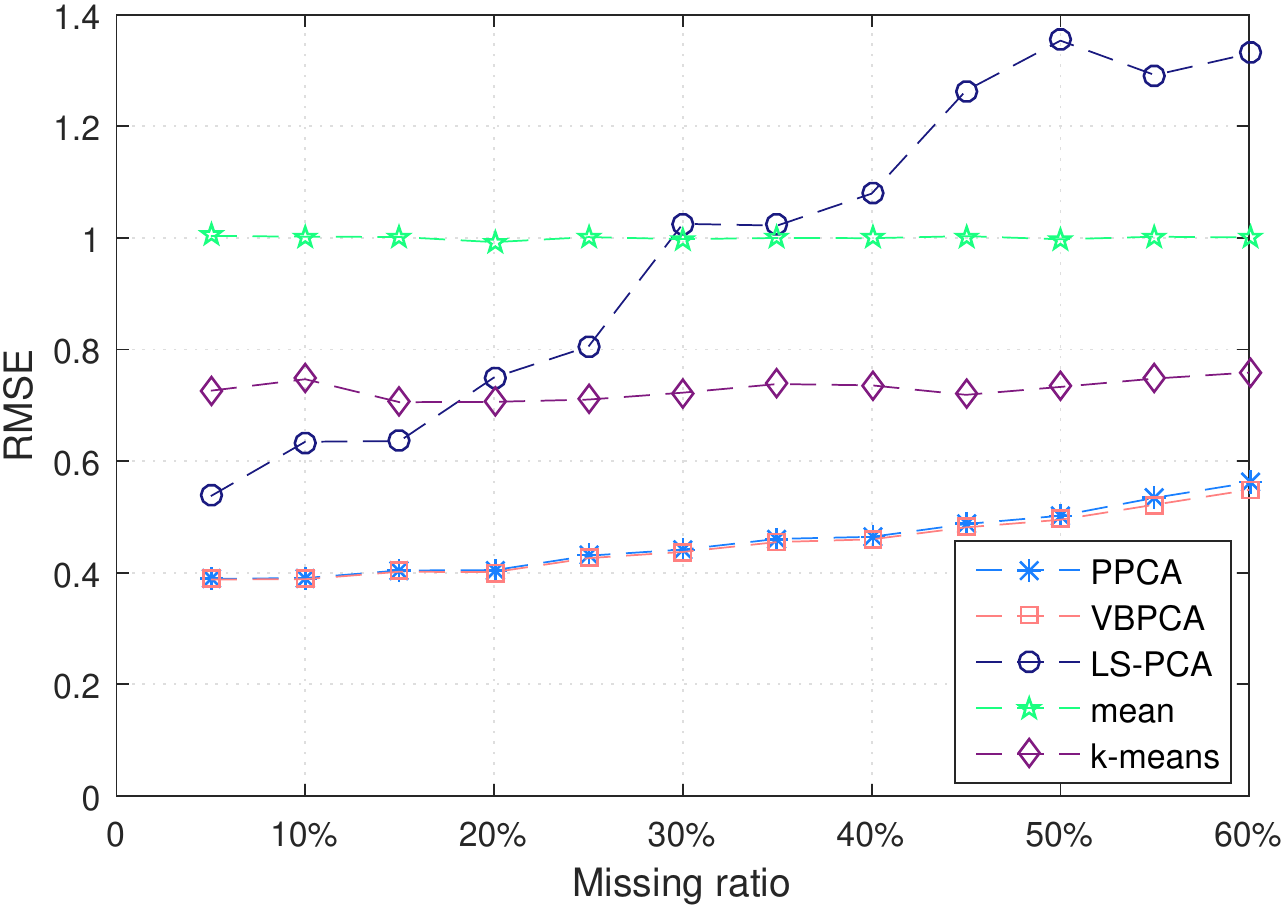}
	\caption{Comparison of imputation algorithms.}
	\label{fig5}
\end{figure}

As shown in Fig. \ref{fig5}, the results of \textit{Experiment II} show that RMSEs of PPCA and VBPCA are comparable and much lower than that of the mean imputation and $k$-means imputation under any missing ratios. It is not surprising that RMSE of LS-PCA increases rapidly with the the missing ratio, since the overfitting problem of LS-PCA becomes more severe under a high missing ratio. RMSE of the mean imputation remains stable around 1, because the explanatory variable values in the dataset are standardized, with a mean value of 0 and standard deviation of 1.

%%%%%%%%%%%%%%%%%%%%%%%%%%%%%%%%
%%%%%%%%%%%%%%%%%%%%%%%%%%%%%%%%

\subsection{Sensitive Analysis of Missing Data on Predictive Performance}

In this section, the sensitivity analysis is implemented to see how the predictive performance of the classifiers (including SVM with linear, Gaussian, and polynomial kernels and AdaBoost) react with the increase of missing ratios.

\begin{figure}[!ht]
	\centering
	\includegraphics[width=0.9\linewidth]{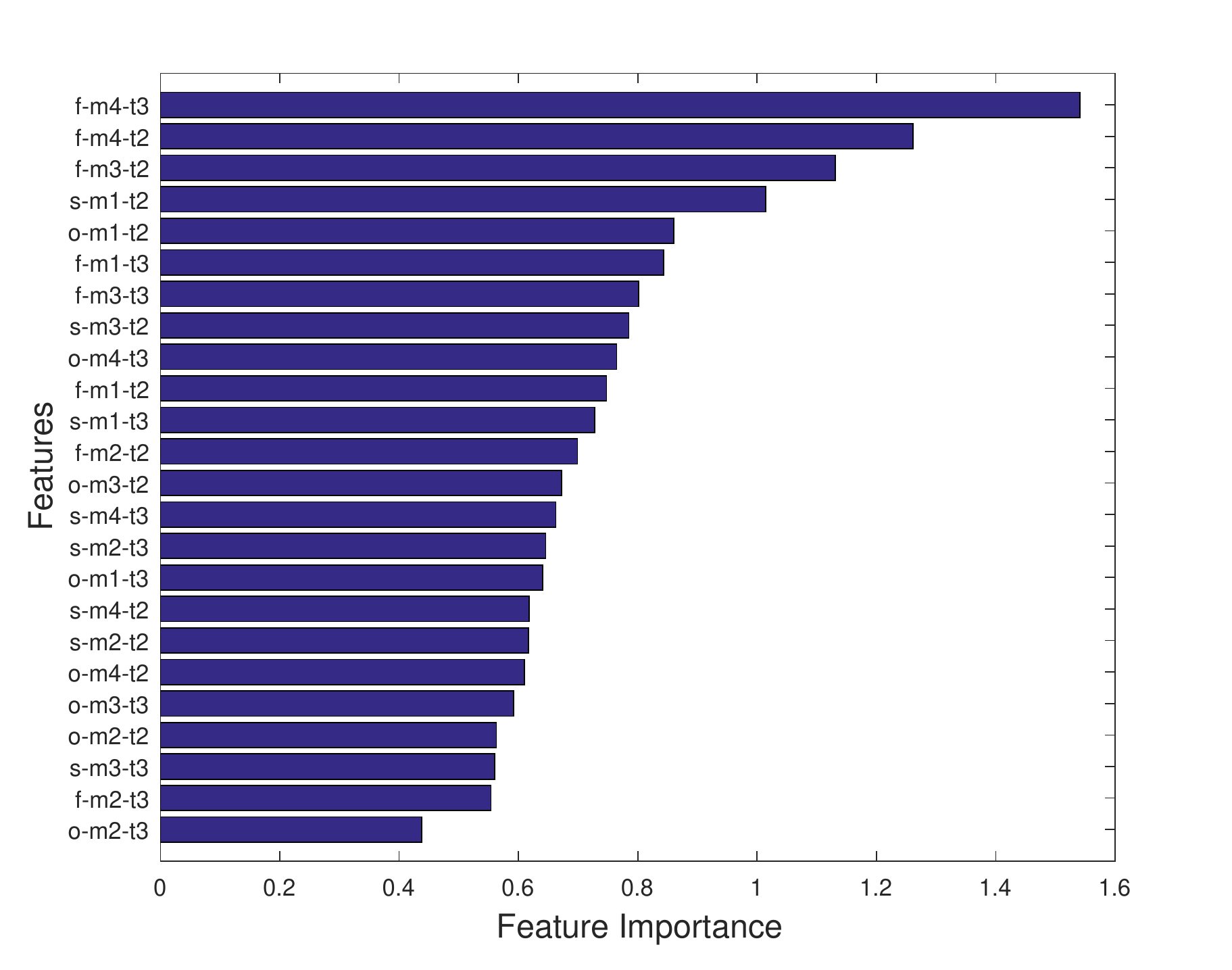}
	\caption{Feature importance ranking by the random forest.}
	\label{fig6}
\end{figure}

Firstly, the random forest is utilized to calculate the feature importance of the explanatory variables (see Fig. \ref{fig6}). It is observed that the flow rate 10-15 min ahead of crash at the second downstream RTMS ($f-m4-t3$), the flow rate 5-10 min ahead of crash at the second downstream RTMS (f-m4-t2), the flow rate 5-10 min ahead of crash at the first downstream RTMS (f-m3-t2) are the top 3 most important features. Feature selection is proved to be an efficient method to avoid overfitting issue. In this paper, we compare two cases for each classifier: (I) no feature selection is applied, indicating that all the features are used (full features); (II) the top 8 most important features are selected for forecasting (selected features). Therefore, ten predictive models are examined and compared (see Table \ref{table:models}).

\begin{table*}[!t]
	\caption{Ten predictive models}
	\centering
	\footnotesize
	\label{table:models}
	\begin{tabularx}{1.05\textwidth}{lX}
		\toprule
		\toprule
		Model & Description \\
		\midrule
		SVM-linear (full features) & SVM with linear kernel (without feature selection) \\
		SVM-linear (selected features) & SVM with linear kernel (with feature selection) \\
		SVM-Gaussian (full features) & SVM with Gaussian kernel (without feature selection) \\
		SVM-Gaussian (selected features) & SVM with Gaussian kernel (with feature selection) \\
		SVM-polynomial (full features) & SVM with polynomial kernel (without feature selection) \\
		SVM-polynomial (selected features) & SVM with polynomial kernel (with feature selection) \\
		AdaBoost (full features) & AdaBoost (without feature selection) \\
		AdaBoost (selected features) & AdaBoost (with feature selection) \\
		\bottomrule
		\bottomrule
	\end{tabularx}
\end{table*}

\begin{figure}[!ht]
	\centering
	\captionsetup[sub]{font=large,labelfont={bf,sf}}
	\subfloat[SVM-linear (full features)]{\includegraphics[width=0.5\linewidth]{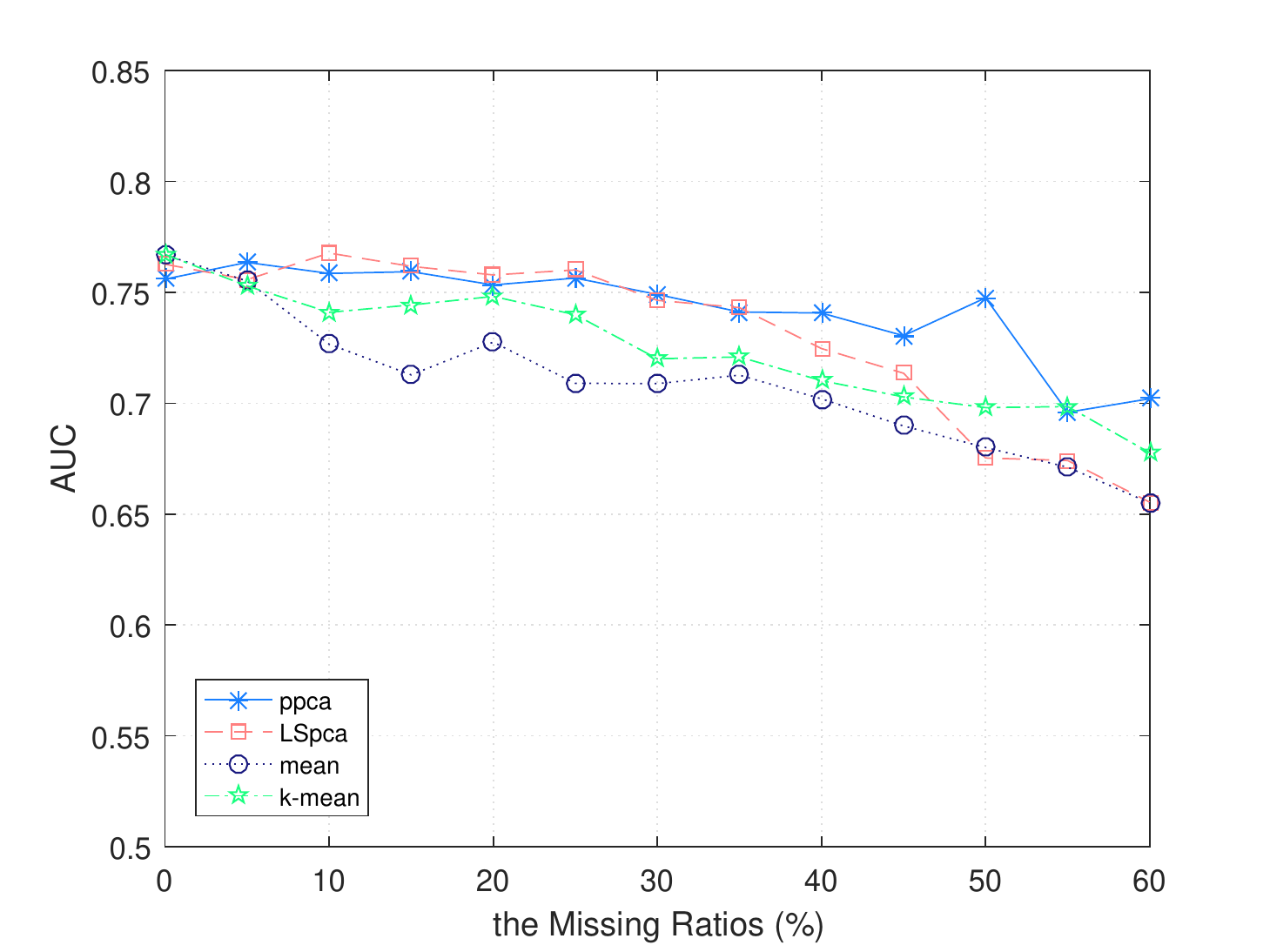}%
		\label{fig7a}}
	\hfil
	\subfloat[SVM-linear (selected features)]{\includegraphics[width=0.5\linewidth]{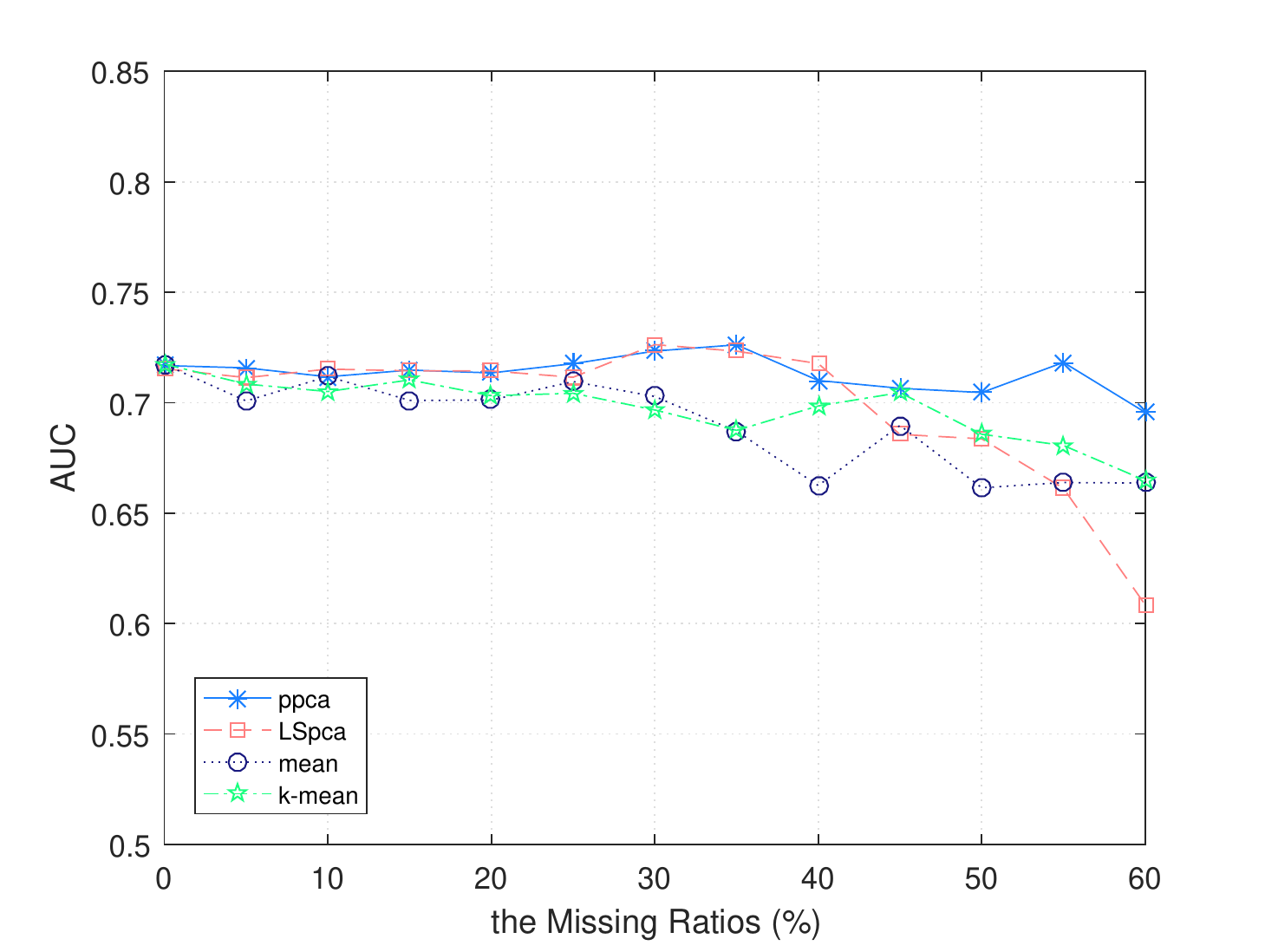}%
		\label{fig7b}}
	\\
	\subfloat[SVM-Gaussian (full features)]{\includegraphics[width=0.5\linewidth]{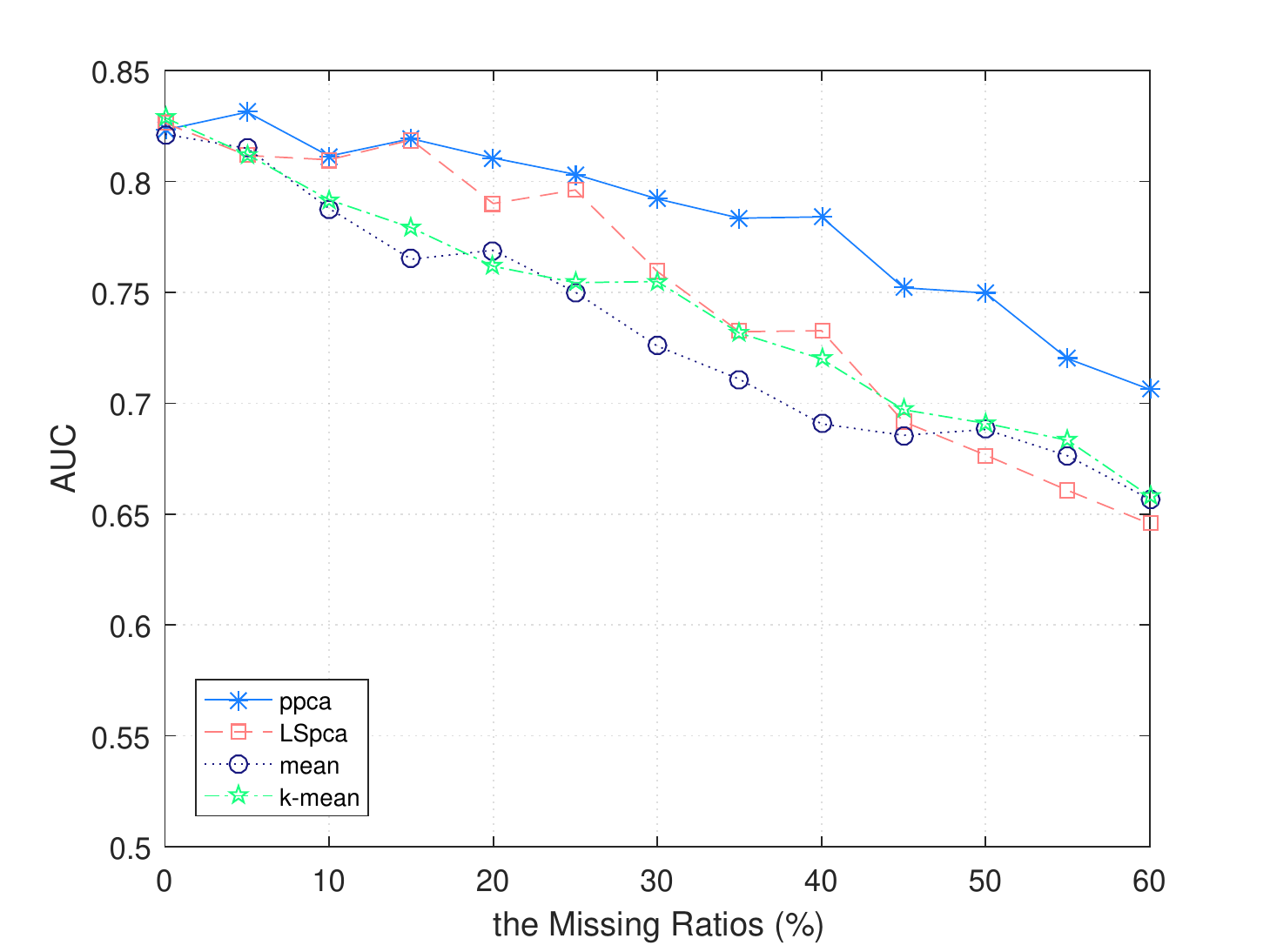}%
		\label{fig7c}}
	\hfil
	\subfloat[SVM-Gaussian (selected features)]{\includegraphics[width=0.5\linewidth]{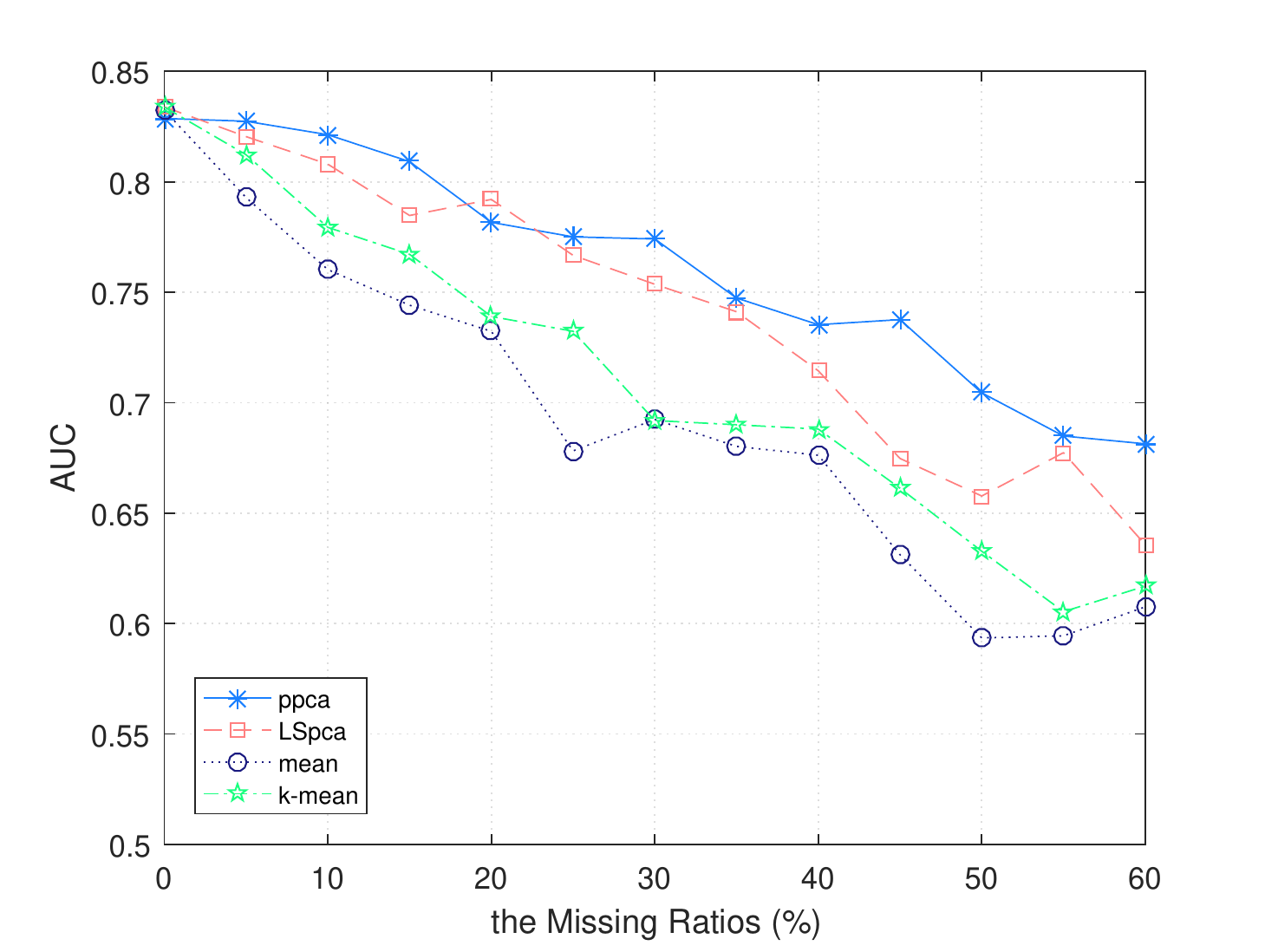}%
		\label{fig7d}}
	\phantomcaption
\end{figure}

\begin{figure}[!ht]
	\ContinuedFloat
	\centering
	\captionsetup[sub]{font=large,labelfont={bf,sf}}
	\subfloat[SVM-polynomial (full features)]{\includegraphics[width=0.5\linewidth]{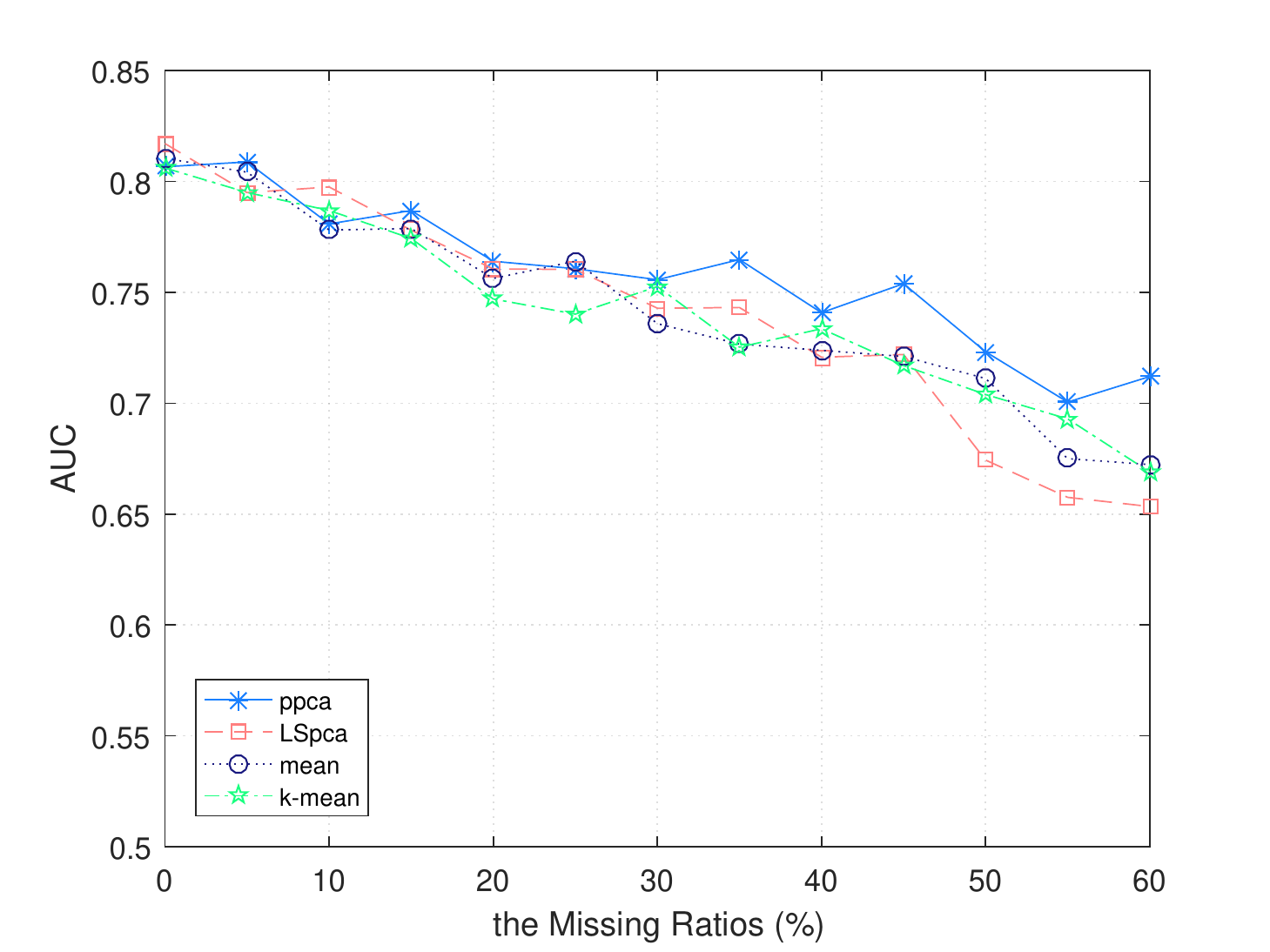}%
		\label{fig7e}}
	\hfil
	\subfloat[SVM-polynomial (selected features)]{\includegraphics[width=0.5\linewidth]{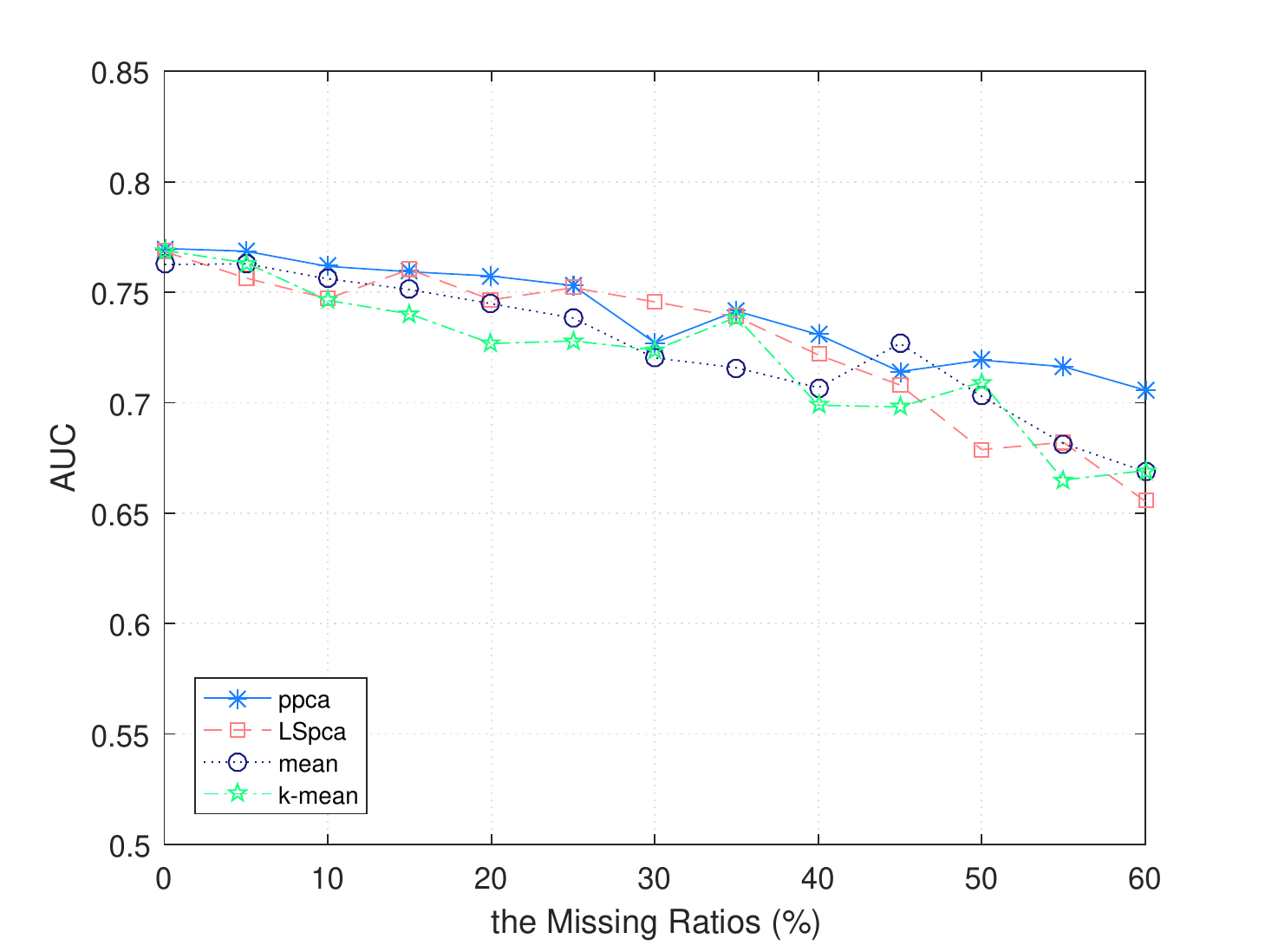}%
		\label{fig7f}}
	\\
	\subfloat[AdaBoost (full features)]{\includegraphics[width=0.5\linewidth]{fig7g}%
		\label{fig7g}}
	\hfil
	\subfloat[AdaBoost (selected features)]{\includegraphics[width=0.5\linewidth]{fig7h}%
		\label{fig7h}}
	\caption{AUCs of 8 classifiers under different missing ratios}
	\label{fig7}
\end{figure}

Secondly, several fractions (ranging from 0 to 60\%, with a step of 5\%) of data are removed from the train-and-test dataset, based on the MCAR pattern. Four kinds of missing data imputation algorithms (i.e., PPCA, LS-PCA, mean imputation, and $k$-means clustering imputation) are utilized to impute the missing values. The previous result shows that PPCA and VBPCA are comparable in terms of the imputation accuracy, thus only PPCA is examined in this section.

Finally, the 8 predictive models are trained and tested (with 10-fold cross validation) on the dataset with different missing ratios, and the AUCs, which measures the overall predictive performance, are calculated. Based on the study in Section 4.2, both COST and SMOTE perform well in solving the imbalanced issues, and they do not show a significant distinction. Therefore, we do not compare different solutions to the imbalanced dataset in this section, but simply select COST as a demonstration for all predictive models. To ensure reliability, all the trails are repeated by five times and the averaged AUCs are recorded.

As shown in Fig. \ref{fig7}, the PPCA imputation outperforms the other three imputation approaches in terms of AUC. The AUCs of the classifiers utilizing the LS-PCA imputation show the fastest decrease with the increase of the missing ratio, which is consistent with the accuracy analysis of imputation approaches in Section 4.3. It is found that the PPCA shows a greater power under higher missing ratios, where the difference between models utilizing PPCA and models using other imputations is significant. This indicates that PPCA has a stable ability to recover the missing information even in a highly missing dataset.

It is observed that the SVM (with Gaussian or polynomial kernels) and the AdaBoost with full features achieve AUCs higher than 0.8, while the AUC of SVM with the linear kernel only reaches 0.76, under complete dataset. However, with the increase of the missing ratios, SVM with the polynomial kernel show a sharp decrease in AUC, while the SVM with the linear kernel is slightly affected. It is also interesting to find that the AUCs of the models except SVM with the polynomial kernel converge to around 0.7 at a high missing ratio (such as 0.6) when the PPCA imputation is used. On the other hand, feature selection can not significantly improve the classifiers' predictive performance in the train-and-test dataset, which indicates the gains from reducing overfitting do not exceed the losses of discarded information.

These results and discussions might provide suggestions to traffic managers on how to select suitable classifiers and imputation approaches in real-time crash likelihood prediction:

\begin{enumerate}[(I)]
	\item It is highly suggested to use PPCA or VBPCA for the missing data imputation, especially in the dataset with high missing ratios.
	\item In the case that the missing ratio is low, it is suggested to use SVM with Gaussian and polynomial kernels, instead of SVM with the linear kernel. However, when the missing ratio is high, SVM with the polynomial kernel should be avoided since its AUC drops dramatically, while other classifiers, such as SVM with linear and Gaussian kernels, and AdaBoost, can achieve comparable AUCs.
\end{enumerate}

\subsection{Out-of-Experiment Validation}

\begin{table}[!t]
	\caption{MoEs in the validation dataset}
	\label{table:validation}
	\centering
	\small
	\begin{tabularx}{0.8\textwidth}{ccccc}
		\toprule
		\toprule
		classifier & accuracy & AUC & sensitivity & specificity \\
		\midrule
		SVM-linear & 0.755 & 0.763 & 0.625 & 0.768 \\
		SVM-Gaussian & 0.736 & 0.742 & 0.575 & 0.752 \\
		SVM-polynomial & 0.740 & 0.740 & 0.550 & 0.759 \\
		AdaBoost & 0.733 & 0.758 & 0.600 & 0.747 \\
		\bottomrule
		\bottomrule
	\end{tabularx}
\end{table}

A series of sensitivity analyses have been conducted in the train-and-test dataset, while an out-of-experiment validation is carried out in the validation dataset. In this validation, the PPCA approach is applied as the missing data imputation method, COST is used as solutions to the imbalanced dataset, while the aforementioned five classifiers are examined.

The results in Table \ref{table:validation} show that the five classifiers achieve AUCs around 0.75 in the validation dataset with 21\% missing data in the real world, which are comparable to the results in the train-and-test dataset. The AUCs are sightly lower than those in the previous studies which were based on the complete dataset, the AUCs of which were roughly in the range between 0.75 and 0.8 \citep{yu2013utilizing,xu2014using}. In this validation dataset, SVM with the linear kernel has the best performance, with the AUC of 0.763, sensitivity of 0.625, specificity of 0.768, and accuracy of 0.755.

\section{Conclusions}

This paper attempts to address the missing data imputation problem and solutions to the imbalanced dataset in real-time crash likelihood estimation. Although these two problems are easily encountered in real-world applications, few research has been conducted in the domain of real-time crash likelihood estimation.

In terms of the missing data imputation, we compare PCA-based missing data imputation algorithms (including PPCA, VBPCA, and LS-PCA) with several conventional imputing approaches, such as the mean imputation and $k$-means clustering imputation. Numerical results show that PPCA and VBPCA outperform LS-PCA and the conventional imputing methods in terms of RMSE, and also help the classifiers achieve more stable predictive performance under high missing ratios. It is found that the two solutions, i.e., cost-sensitive learning techniques on the algorithmic level and SMOTE on the data level, both achieve good performance in improving the sensitivity with an acceptable loss of specificity by adjusting the classifiers to pay more attention to the crash cases (the minority class in the dataset).

It is observed that different classifiers have different decreasing curves measured by AUC with the increase of missing ratios in the train-and-test dataset. SVM with the linear kernel is the weakest classifier when the dataset is complete, but its predictive performance drops more slowly than other classifiers. SVM with the polynomial kernel has outstanding performance in the complete dataset but becomes the worst one under high missing ratios.

An out-of-experiment validation is implemented using an independent dataset, which has 21\% of missing data and is also imbalanced (the ratio of the number of non-crash samples to crash samples is 10:1). In such a partly-missing and imbalanced dataset, the classifiers can achieve AUCs around 0.75, with the help the PPCA missing data imputation and cost-sensitive learning technique.

These interesting findings provide useful insights for the traffic operators to implement proper predictive strategies:

\begin{enumerate}[(I)]
	\item PPCA and VBPCA are two highly-suggested missing data imputing approaches, especially when the missing ratio is high;
	\item The cost-sensitive learning technique and SMOTE are useful and comparable methods to deal with imbalanced issues;
	\item Complex and high-dimensional models, like SVM with the polynomial kernel, are not always the most accurate and stable classifier, especially when the missing ratio is high. The operators should select different classifiers under different circumstances (for example, considering the missing ratio of the real-time traffic flow data).
\end{enumerate}

The limitations and potential future are discussed as follows. Firstly, only the MCAR case is considered in this paper, while the missing pattern of NMAR is also commonly observed in field applications (for example, the failure of sensors could lead to a long and continuous missing sequences in traffic flow data). In such scenarios, the case-by-case design should be implemented for missing data imputation. In the future, we expect to propose approaches to dealing with NMAR patterns, and explore more missing data imputation algorithms, such as the tensor decomposition, in the domain of real-time crash likelihood estimation. Secondly, this paper only considers the missing features of crash/non-crash samples, but does not tackle the issue of missed crash records. The issue of missing crash samples is more difficult than that of missing features, since it makes the classification to be a semi-supervised learning problem. We hope to investigate this problem in our future studies.

\section*{Acknowledgements}
This research is financially supported by Zhejiang Provincial Natural Science Foundation of China [LR17E080002], National Natural Science Foundation of China [51508505, 71771198, 51338008], and Fundamental Research Funds for the Central Universities [2017QNA4025].

\section*{References}

\bibliography{articles}

\end{document}